\documentclass[sigconf]{acmart}
\usepackage{tikz}
\usepackage{standalone}
\DeclareUnicodeCharacter{0307}{.}

\AtBeginDocument{%
  }

\setcopyright{acmlicensed}
\copyrightyear{2025}
\acmYear{2025}
\acmDOI{XXXXXXX.XXXXXXX}
\acmConference[KDD 2025]{SIGKDD International Conference on Knowledge Discovery and Data Mining 2025}{August 03-07,
  2025}{Toronto, Canada}
\acmISBN{978-1-4503-XXXX-X/2018/06}

\usepackage{multirow}

\copyrightyear{2025}
\acmYear{2025}
\setcopyright{acmlicensed}\acmConference[KDD '25]{Proceedings of the 31st ACM SIGKDD Conference on Knowledge Discovery and Data Mining V.2}{August 3--7, 2025}{Toronto, ON, Canada}
\acmBooktitle{Proceedings of the 31st ACM SIGKDD Conference on Knowledge Discovery and Data Mining V.2 (KDD '25), August 3--7, 2025, Toronto, ON, Canada}
\acmDOI{10.1145/3711896.3736558}
\acmISBN{979-8-4007-1454-2/2025/08}

\begin{document}

\title{Relational Deep Learning: Challenges, Foundations and Next-Generation Architectures}



\author{Vijay Prakash Dwivedi}
\affiliation{%
  \institution{Stanford University}
  \city{Stanford}
  \state{California}
  \country{USA}
}
\email{vdwivedi@cs.stanford.edu}

\author{Charilaos Kanatsoulis}
\affiliation{%
  \institution{Stanford University}
  \city{Stanford}
  \state{California}
  \country{USA}
}
\email{charilaos@cs.stanford.edu}

\author{Shenyang Huang}
\affiliation{%
  \institution{Mila, McGill University}
  \city{Montreal}
  \state{Quebec}
  \country{Canada}
}
\email{shenyang.huang@mail.mcgill.ca}

\author{Jure Leskovec}
\affiliation{%
  \institution{Stanford University}
  \city{Stanford}
  \state{California}
  \country{USA}
}
\email{jure@cs.stanford.edu}





\renewcommand{\shortauthors}{Vijay Prakash Dwivedi, Charilaos Kanatsoulis, Shenyang Huang, \& Jure Leskovec}

\begin{abstract}

  Graph machine learning has led to a significant increase in the capabilities of models that learn on arbitrary graph-structured data and has been applied to molecules, social networks, recommendation systems, and transportation, among other domains. Data in multi-tabular relational databases can also be constructed as `relational entity graphs’ for Relational Deep Learning (RDL) - a new blueprint that enables end-to-end representation learning without traditional feature engineering. Compared to arbitrary graph-structured data, relational entity graphs have key properties: (i) their structure is defined by primary-foreign key relationships between entities in different tables, (ii) the structural connectivity is a function of the relational schema defining a database, and (iii) the graph connectivity is temporal and heterogeneous in nature. In this paper, we provide a comprehensive review of RDL by first introducing the representation of relational databases as relational entity graphs, and then reviewing public benchmark datasets that have been used to develop and evaluate recent GNN-based RDL models. We discuss key challenges including large-scale multi-table integration and the complexities of modeling temporal dynamics and heterogeneous data, while also surveying foundational neural network methods and recent architectural advances specialized for relational entity graphs. Finally, we explore opportunities to unify these distinct modeling challenges, highlighting how RDL converges multiple sub-fields in graph machine learning towards the design of foundation models that can transform the processing of relational data.
\end{abstract}

\newcommand{\ah}[1]{{{\textcolor{blue}{\textbf{AH: }}}{\textcolor{blue}{#1}}}}
\newcommand{\ck}[1]{{{\textcolor{purple}
{\textbf{CK: }}}{\textcolor{purple}{#1}}}}
\newcommand{\vd}[1]{{{\textcolor{olive}
{\textbf{VD: }}}{\textcolor{olive}{#1}}}}
\newcommand{\jl}[1]{{{\textcolor{red}
{\textbf{JL: }}}{\textcolor{olive}{#1}}}}




\settopmatter{printfolios=true}

\maketitle
\section{Introduction}
\label{sec:intro}

Graph Neural Networks (GNNs) \cite{kipf2016semi, hamilton2017inductive, gilmer2017neural, velickovic2017graph} have transformed the way we process and extract predictive signals from arbitrary graph-structured data. As a core class of methods in graph machine learning, GNNs receive input graph datasets that include features on nodes and edges along with the connectivity patterns among nodes, and they learn feature representations in an end-to-end manner using modern deep learning techniques. Compared to traditional feature engineering on graph datasets \cite{gallagher2008leveraging, henderson2012rolx, grover2016node2vec}, GNNs allow automatic feature representation learning through deep neural networks' approach \cite{lecun2015deep} and enable the scaling of predictive tasks in industrial applications with large-scale graphs such as social-network recommendations \cite{ying2018graph, zhao2025gigl}, retail \cite{jainenhancing, yang2019aligraph}, transportation \cite{derrow2021eta}, and molecular prediction \cite{sypetkowski2025scalability}.

\textbf{Traditional ML on databases.}
Relational databases (RDBs) store much of the world’s structured data across multiple tables, each corresponding to a different entity type defined by column values. These entities are connected across tables through primary and foreign key relationships, allowing the databases to be represented as relational entity graphs, where a node denotes an entity in a table and an edge denotes the connection between two entities in two tables defined by primary and foreign keys. Such graphs can be large-scale, comprising millions of entities or more and their connections. 
Traditionally, prediction tasks on relational databases have been performed in two steps: first, a feature engineering phase that composes data from different tables using SQL operations and similar techniques, and second, standard model training using methods such as XGBoost \cite{chen2016xgboost} or other tabular models. However, these conventional approaches fail to leverage the rich structural information that binds entities together in a database.

\begin{figure}[t]
    \begin{center}
    \includegraphics[width=0.9\columnwidth]{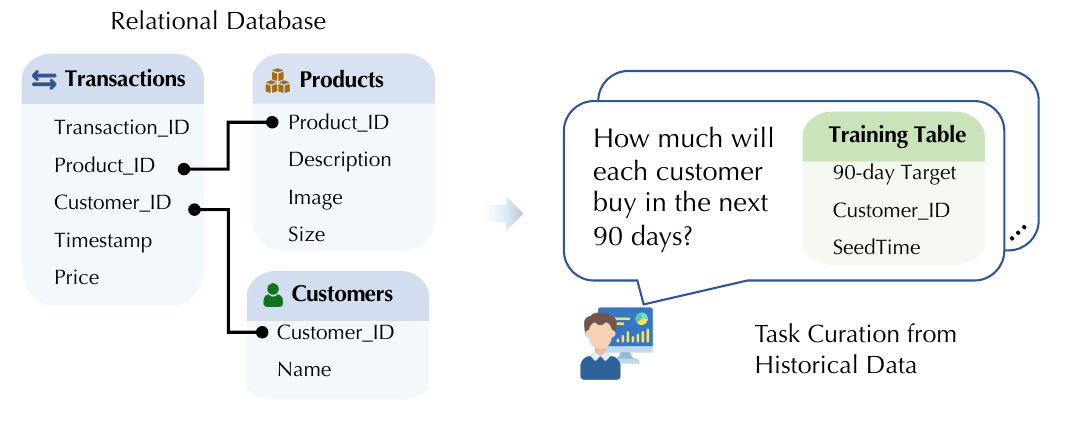}
    \vspace{-10pt}
        \caption{Illustration of a relational database with an example task of predicting the value a customer will purchase in the next 90 days. The details related to the task is curated into a training table by collecting historical data from the database. Figure adapted from \citet{fey2023relational}.} 
        \Description[Illustration of a relational database]{Illustration of a relational database with an example task of predicting the value a customer will purchase in the next 90 days. The details related to the task is curated into a training table by collecting historical data from the database.}
        \vspace{-10pt}
        \label{fig:intro}
    \end{center}
\end{figure}

\textbf{Relational deep learning.}
In contrast, recent advances in graph deep learning have demonstrated that neural networks can directly utilize graph structural information to achieve state-of-the-art performance. Recently, \citet{fey2023relational} introduced a blueprint called `Relational Deep Learning (RDL)' that enables the direct use of relational entity graphs from relational databases in graph neural networks, thereby alleviating the need for the traditional two-stage machine learning process that is bottlenecked by conventional feature engineering.
In RDL, as illustrated in Figure \ref{fig:intro}, a relational database is represented as a relational entity graph \cite{fey2023relational} - typically on a larger scale (often comprising millions of nodes) compared to standard graph benchmarks. For example, a user in \texttt{Customers} table of a relational database can be connected to an item in \texttt{Products} table through a transaction in \texttt{Transactions} table which also records the timestamp of the transaction. Observe that there are three node types in this graph snippet that corresponds to the database, and the same user may have multiple transactions at different times. Thus, RDL naturally represents relational tables as graphs that are temporal, heterogeneous, and large-scale, in contrast to most graph machine learning benchmarks and methods that focus on static, homogeneous graphs of medium-to-small scale.

To develop graph neural network based methods on such relational entity graphs, \citet{robinson2025relbench} developed a RDL benchmark-RelBench-that brings together relational entity graphs from diverse sources such as e-commerce, social networks, sports and medical domains, under a unified framework with standard data splits, training and evaluation pipelines and an open-source repository with leaderboard to track progress of developments in this field. GNN based RDL models outperform (or achieve comparable performance than) traditional feature engineering based techniques, while significantly reducing the total model development time (by over 95\%) \cite{robinson2025relbench}. The models employed in \citet{robinson2025relbench} use heterogeneous GraphSAGE \cite{hamilton2017inductive} networks that depends on a temporal-aware graph sampling.

\textbf{Recent advances.}
Recently, several works have emerged in this sub-field of graph machine learning, focusing on developing improved message-passing for heterogeneous graph learning \cite{ferrini2024self, chen2025relgnn, pelevska2024transformers}, leveraging retrieval-augmented generation (RAG) techniques with large language models \cite{wydmuch2024tackling}, and employing hybrid GNN and tabular methods \cite{lachiover}, among others. While these approaches are promising, we believe that the success of RDL depends on the convergence of several sub-fields in graph machine learning - such as modeling temporal dynamics in graphs, scaling methods to large-scale data, developing unified modeling techniques for multiple domains, extending homogeneous graph methods to heterogeneous settings, creating Transformers for graphs, and ultimately establishing foundation models for graphs. However, many of these subtopics have segregated benchmarks and lack a unified path for progress. For instance, works proposing Transformers for graphs often do not address their direct application to temporal and heterogeneous graphs \cite{rampavsek2022recipe}. Similarly, graph foundation models developed for the molecular domain are inherently non-transferable to other application areas \cite{sypetkowski2025scalability}, and nodes and edges are defined inconsistently across domains \cite{hu2020open}. For these reasons, RDL represents a converging paradigm for several sub-fields in graph machine learning, while simultaneously posing the challenge of unifying different aspects to push the boundaries of graph machine learning and enable powerful end-to-end deep learning on relational databases.

\textbf{Present work.} In this paper, we provide a comprehensive review of the challenges of relational deep learning by reviewing its formulation and multiple data modeling aspects, in Section \ref{sec:datasets}; revisiting foundational methods in relational and graph learning such as tabular learning models, GNNs, temporal and heterogeneous GNNs and Graph Transformers, in Section \ref{sec:methods}; showing the recent progress in models targeting RDL as well as the frontier developments in other graph modeling aspects such as temporality and heterogeneity, in Section \ref{sec:frontiers}. Finally, we outline future directions in Section \ref{sec:future} that build on the current advances and aim to facilitate the convergence of these diverse sub-topics in graph machine learning.

\section{Relational Deep Learning}
\label{sec:datasets}

In this section, we provide an introduction of Relational Deep Learning, discuss the natural transformation of databases to graphs, review public benchmark datasets and finally outline its challenges.

\begin{figure*}[t]
    \begin{center}
    \includegraphics[width=0.7\textwidth]{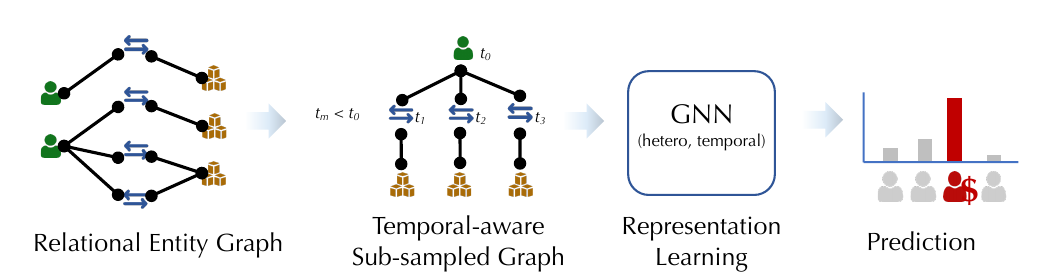}
        \caption{Overview of the Relational Deep Learning (RDL) pipeline: The relational entity graphs are first processed by a temporal-aware sampler, which extracts relevant subgraphs for the seed nodes which have a seed time for its prediction. These inputs are then passed through a graph neural network (GNN) that captures both the heterogeneity and temporal dynamics of the graph, enabling end-to-end representation learning for target label prediction.} 
        \Description[Overview of the RDL pipeline]{Overview of the Relational Deep Learning (RDL) pipeline: The relational entity graphs are first processed by a temporal-aware sampler, which extracts relevant subgraphs for the seed nodes which have a seed time for its prediction. These inputs are then passed through a graph neural network (GNN) that captures both the heterogeneity and temporal dynamics of the graph, enabling end-to-end representation learning for target label prediction.}
        \label{fig:pipeline}
    \end{center}
\end{figure*}

\subsection{Relational Databases} \label{sub:database}
A relational database, which can be represented by $(\mathcal{T}, \mathcal{L})$, consists of multiple tables $\mathcal{T}=\{T_1, \dots, T_n\}$ and inter-table links $\mathcal{L}\subseteq\mathcal{T}\times\mathcal{T}$. A link $L=(T_{\text{fkey}},T_{\text{pkey}})\in\mathcal{L}$ exists if a foreign key in table $T_{\text{fkey}}$ references a primary key in table $T_{\text{pkey}}$. Each table $T$ is a set of rows (entities) $\{v_1,\dots,v_{n_T}\}$. 
An entity $v\in T$ consists of four components: 
\begin{enumerate}
    \item \emph{a primary key} $p_v$ which uniquely identifies entity $v$.
    \item \emph{foreign keys} $K_v \subseteq \{p_{v'} : v' \in T', (T,T')\in\mathcal{L}\}$, which reference or link primary keys in related tables.
    \item \emph{attributes} $x_v=(x_v^1,\dots,x_v^{d_T})$, which contain entity-specific data, the design of which is shared by entities within a table.
    \item \emph{a timestamp} $t_v$ which records the occurrence time of events or in particular the entity.
\end{enumerate}

An example of a relational database is illustrated in Fig. \ref{fig:intro}, which comprises 3 tables: \texttt{Transactions}, \texttt{Products}, and \texttt{Customers}, with their corresponding attribute schema included. Each table encodes entities as rows. 
For instance, every row $v$ in \texttt{Transactions} is uniquely identified by (i). \emph{a primary key} \(p_v = \texttt{Transaction\_ID}\) and includes (ii). \emph{foreign keys}
\(K_v\) such as \texttt{Product\_ID} and \texttt{Customer\_ID}, referencing the primary keys of \texttt{Products} and \texttt{Customers}, respectively, (iii) \emph{attributes} \(x_v = (x_v^1, \dots, x_v^{d_T})\) such as \texttt{Price}, and (iv) \emph{a timestamp} \(t_v = \texttt{Timestamp}\) indicating when the transaction occurred. Similarly, rows in \texttt{Products} and \texttt{Customers} tables have their respective primary keys, and may hold attributes such as \texttt{Description} or \texttt{Name} among others. As in the figure, the foreign key references in \texttt{Transactions} point to the corresponding entities in \texttt{Products} and \texttt{Customers}, thereby forming the inter-table links needed to build the relational entity graph, which forms the input for the RDL pipeline, further introduced in Figure \ref{fig:pipeline}.

\subsection{Relational Entity Graphs} \label{sub:rel_graph}
As observed from Figure \ref{fig:intro}, a relational database (RDB) naturally forms a graph-structured dataset. In this graph, nodes correspond to entities from different tables $\mathcal{T}$, and edges represent the primary-foreign key relationships linking these entities. The graph is heterogeneous, as each node type is associated with a distinct table, and temporal, since timestamps are recorded for certain nodes (e.g., \texttt{Transactions} as shown in Figure \ref{fig:intro}). This graph structured data is referred to as a relational entity graph in the RDL blueprint \cite{fey2023relational}.

\begin{definition}[Relational Entity Graph]
A Relational Entity Graph is defined as a heterogeneous, temporal graph $\mathcal{G} = (\mathcal{V}, \mathcal{E}, \phi, \psi, \tau)$, where $\mathcal{V}$ is the set of nodes (each corresponding to an entity from a table in $\mathcal{T})$, $\mathcal{E}$ is the set of edges (representing the primary-foreign key links), $\phi: \mathcal{V} \rightarrow \mathcal{A}$ maps each node to a type determined by its originating table, $\psi: \mathcal{E} \rightarrow \mathcal{R}$ assigns each edge a relation type according to its connection, and $\tau: \mathcal{V} \cup \mathcal{E} \rightarrow \mathbb{R}$ assigns timestamps that capture temporal properties such as transaction times.
\end{definition}
The input node features are derived from the values in the columns of a table corresponding to each row (node). For every \(v \in \mathcal{V}\), the feature is defined as \(\mathbf{h}_v \in \mathbb{R}^{d_{\phi(v)}}\), where \(d_{\phi(v)}\) denotes the dimensionality of the feature space for node type \(\phi(v)\). These embeddings are generated via PyTorch Frame \cite{hu2024pytorch}, which utilizes learnable multimodal column encoders to transform raw column data into vector representations. Consequently, an RDB can be represented as a large-scale, temporal, and heterogeneous relational entity graph, potentially comprising millions of nodes corresponding to the rows across its various tables. These graphs can be fed to an end-to-end GNN 
pipeline and can be applied to downstream prediction tasks, as illustrated in Figure \ref{fig:pipeline}.

\begin{table*}[t]
\centering
\scalebox{0.75}{
\begin{tabular}{lclrrrlc}
\toprule
\textbf{Name} & \textbf{Domain} & \textbf{\#Tasks (Types)} & \textbf{\#Tables} & 
\textbf{\#Rows} & \textbf{\#Cols} & \textbf{Source} & \textbf{Reference}\\
\midrule
\texttt{rel-amazon} & E-commerce & 7 (\texttt{clf}, \texttt{reg}, \texttt{rec}) & 3  & 15,000,713 & 15 & \url{https://relbench.stanford.edu/datasets/rel-amazon} & \multirow{7}{*}{\cite{robinson2025relbench}}\\
\texttt{rel-avito}  & E-commerce & 4 (\texttt{clf}, \texttt{reg}, \texttt{rec})  & 8  & 20,679,117 & 42 & \url{https://relbench.stanford.edu/datasets/rel-avito} & \\
\texttt{rel-event}  & Social     & 3 (\texttt{clf}, \texttt{reg})  & 5  & 41,328,337 & 128 & \url{https://relbench.stanford.edu/datasets/rel-event} & \\
\texttt{rel-f1}     & Sports     & 3 (\texttt{clf}, \texttt{reg}) & 9  & 74,063     & 67 & \url{https://relbench.stanford.edu/datasets/rel-f1} & \\
\texttt{rel-hm}     & E-commerce & 3 (\texttt{clf}, \texttt{reg}, \texttt{rec})  & 3  & 16,664,809 & 37 & \url{https://relbench.stanford.edu/datasets/rel-hm} & \\
\texttt{rel-stack}  & Social     & 5 (\texttt{clf}, \texttt{reg}, \texttt{rec})  & 7  & 4,247,264  & 52 & \url{https://relbench.stanford.edu/datasets/rel-stack} &  \\
\texttt{rel-trial}  & Medical    & 5 (\texttt{clf}, \texttt{reg}, \texttt{rec})  & 15 & 5,434,924  & 140 & \url{https://relbench.stanford.edu/datasets/rel-trial} &  \\
\midrule
\texttt{accidents} & Government & 1 (\texttt{clf}) & 3 & 1,453,650 & 43 & \url{https://relational.fel.cvut.cz/dataset/Accidents} & \multirow{18}{*}{\cite{motl2015ctu}}\\
\texttt{airline} & Retail & 1 (\texttt{clf}) & 19 & 448,156 & 119 & \url{https://relational.fel.cvut.cz/dataset/Airline} & \\
\texttt{basketballmen} & Sports & 1 (\texttt{reg}) & 9 & 43,841 & 195 & \url{https://relational.fel.cvut.cz/dataset/BasketballMen} & \\
\texttt{ccs} & Financial & 1 (\texttt{reg}) & 6 & 422,868 & 29 & \url{https://relational.fel.cvut.cz/dataset/CCS} & \\
\texttt{cdeschools} & Government& 1 (\texttt{reg}) & 3 & 29,481 & 90 & \url{https://relational.fel.cvut.cz/dataset/CDESchools} & \\
\texttt{ergastf1} & Sports& 1 (\texttt{clf}) & 14 & 544,056 & 98 & \url{https://relational.fel.cvut.cz/dataset/ErgastF1} & \\
\texttt{financial} & Financial & 1 (\texttt{clf}) & 8 & 1,090,086 & 55 & \url{https://relational.fel.cvut.cz/dataset/Financial} & \\
\texttt{fnhk} & Medical & 1 (\texttt{reg}) & 3 & 2,108,356 & 24 & \url{https://relational.fel.cvut.cz/dataset/FNHK} & \\
\texttt{geneea} & Government & 1 (\texttt{clf}) & 19 & 799,290 & 128 & \url{https://relational.fel.cvut.cz/dataset/Geneea} & \\
\texttt{legalacts} & Government & 1 (\texttt{clf}) & 5 & 1,756,749 & 33 & \url{https://relational.fel.cvut.cz/dataset/LegalActs} & \\
\texttt{thrombosis} & Medical & 1 (\texttt{clf}) & 3 & 15,899 & 64 & \url{https://relational.fel.cvut.cz/dataset/Thrombosis} & \\
\texttt{monodial} & Geography & 1 (\texttt{clf}) & 40 & 21,497 & 167 & \url{https://relational.fel.cvut.cz/dataset/Monodial} & \\
\texttt{ncaa} & Sports & 1 (\texttt{clf}) & 9 & 201,555 & 106 & \url{https://relational.fel.cvut.cz/dataset/NCAA} & \\
\texttt{premiereleague} & Sports & 1 (\texttt{clf}) & 4 & 10,716 & 217 & \url{https://relational.fel.cvut.cz/dataset/PremiereLeague} & \\
\texttt{seznam} & Retail & 1 (\texttt{reg}) & 4 & 2,681,983 & 14 & \url{https://relational.fel.cvut.cz/dataset/Seznam} & \\
\texttt{sfscores} & Government & 1 (\texttt{reg}) & 3 & 66,153 & 25 & \url{https://relational.fel.cvut.cz/dataset/SFScores} & \\
\texttt{stats} & Education & 1 (\texttt{reg}) & 8 & 1,027,838 & 71 & \url{https://relational.fel.cvut.cz/dataset/Stats} & \\
\texttt{voc} & Retail & 1 (\texttt{clf}) & 8 & 29,067 & 89 & \url{https://relational.fel.cvut.cz/dataset/VOC} & \\
\texttt{walmart} & Retail & 1 (\texttt{reg}) & 4 & 4,628,497 & 27 & \url{https://relational.fel.cvut.cz/dataset/Walmart} & \\
\bottomrule
\end{tabular}
}
\vspace{4pt}
\caption{Overview of the public RDL benchmarks that fulfil the criteria of (i) at least 3 tables, (ii) at least 10,000 rows, (iii) are real, (iv) with temporal records, and (v) a defined prediction task. \texttt{clf}: classification, \texttt{reg}: regression, \texttt{rec}: recommendation.}
\label{tab:datasets-summary}
\vspace{-10pt}
\end{table*}

\subsection{Datasets for RDL} \label{sub:relbench}

\textbf{RelBench.} RelBench \cite{robinson2025relbench} is one of the first benchmarks for relational deep learning. It provides a collection of 7 real-world relational databases which can be represented as a temporal, heterogeneous relational
entity graph, as described in the previous subsections. These databases, as summarized in Table \ref{tab:datasets-summary}, span diverse domains such as e-commerce, social, sports, and medical, range from 3 to 15 tables, and vary in size from 74k to 41m rows.

For each database, RelBench defines a set of predictive tasks, with a total of 30 tasks across 7 databases. The tasks include node-level tasks (e.g., entity classification and regression) and link-level tasks (e.g., recommendation). To construct each task, a `training table' is created from historical data, containing the relevant entities for prediction along with the specific timestamps at which predictions are to be made (as shown in Figure \ref{fig:intro} and \ref{fig:pipeline}). This setup ensures that temporal information is properly incorporated into model training and evaluation. All datasets in RelBench have a temporally defined standard train, val and test split defined with earlier records in train and later records in val and test accordingly.

\textbf{Other Datasets.} In Addition to RelBench, the CTU Relational Dataset Repository \cite{motl2015ctu} maintains a
collection of 83 relational databases from diverse domains, and admit different scales in terms of table number, record rows, and attribute columns. These databases have also been used to evaluate recent methods in multi-tabular deep learning, including GNN- and Transformer-based approaches \cite{zahradnik2023deep, pelevska2024transformers}. Unlike RelBench, the datasets in \citet{motl2015ctu} have a single predefined classification or regression task corresponding to each database. Table \ref{tab:datasets-summary} shows an overview of the repository’s datasets that originate from real-world sources, include temporal records, and contain at least 10,000 rows distributed across three or more tables.

\subsection{Challenges}  \label{sub:challenge}

\textbf{Heterogeneity.} Building effective relational deep learning (RDL) models pose unique challenges because
relational databases can be large, often encompassing millions of rows and
numerous tables \cite{fey2023relational, robinson2025relbench, motl2015ctu}, and have attribute schema varying from one database to another and even within different tables under the same database.
Unlike standard graph learning, these tables store diverse attributes in columns of varying modalities (e.g., text, numerical, images), which must be integrated within a single pipeline. This heterogeneity introduces complexities in data encoding, as well as in ensuring consistent, end-to-end model training that preserves the relational structure of the underlying database.

\textbf{Temporality.} A second challenge lies in effectively handling temporality and preventing time leakage \cite{fey2023relational}. Relational databases frequently record events (e.g., transactions, interaction records) that evolve over time, requiring specialized techniques such as temporal neighbor sampling, as shown in Figure \ref{fig:pipeline}. Standard message passing based GNNs \cite{gilmer2017neural}, which typically assume static or homogeneous graphs, need to be adapted to propagate messages exclusively across time-consistent subgraphs \cite{rossi2020temporal}, ensuring that information from the future has no influence on predictions for earlier timestamps.

\textbf{Graph structure.} Next, relational entity graphs differ from general heterogeneous graphs in a fundamental way. In typical heterogeneous graphs, edge types encode direct semantic interactions between entities. In contrast, relational data graphs are structured around primary-foreign key relationships, which define how tables are connected rather than conveying semantic meaning. This distinction significantly impacts information propagation, underscoring the need for models designed to capture the structural characteristics of relational databases. This difference often restricts the topological patterns available for information flow, highlighting the need for specialized message passing mechanisms \cite{zahradnik2023deep, pelevska2024transformers, chen2025relgnn}.

\textbf{Transfer learning.} Finally, the large-scale and structurally consistent nature of relational entity graphs presents an opportunity for developing foundation models, though it simultaneously poses computational and knowledge transferability challenges \cite{galkin2023towards, mao2024position}. In a manner analogous to how large language models \cite{radford2018improving, raffel2020exploring, tay2022ul2, achiam2023gpt, team2023gemini, bai2023qwen, anthropic2024claude, bi2024deepseek, grattafiori2024llama, team2024reka, team2024gemma} have transformed natural language processing by unifying large corpora of text under a single representation space, foundation models for relational data could similarly reshape how we handle large-scale, multi-table learning. 

\section{Methods}
\label{sec:methods}

\begin{figure*}[t]
    \includegraphics[width=0.7\textwidth]{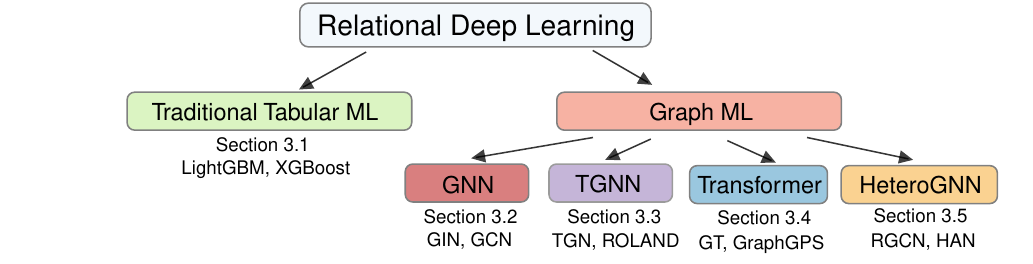}
        \caption{Taxonomy of related areas to Relational Deep Learning.} 
        \Description[Taxonomy of RDL areas.]{Taxonomy of related areas to Relational Deep Learning.}
        \label{fig:taxonomy}
    \vspace{-10pt}
\end{figure*}

Here, we review the methods that can be applied to perform end-to-end representation learning for RDL. Figure~\ref{fig:taxonomy} shows the taxonomy of related areas to RDL. We first provide a background of traditional tabular learning methods and then review graph neural networks (GNNs) in detail, with their temporal extensions, generalization with Transformers, and finally heterogeneous models.

\subsection{Tabular Machine Learning Methods} \label{sub:tab_ml}
Tabular machine learning remains central to industrial applications, with tree-based models such as XGBoost \cite{chen2016xgboost} and LightGBM \cite{ke2017lightgbm} serving as foundational tools due to their efficiency, scalability, and interpretability. At the same time, deep learning approaches tailored for tabular data have gained traction, incorporating techniques like transformers, self-supervised learning, and differentiable decision trees to improve performance \citep{huang2020tabtransformer,arik2021tabnet,gorishniy2021revisiting,gorishniy2022embeddings,chen2023trompt}. While most research in this domain has focused on single-table learning, emerging efforts are beginning to address the challenges of multi-table settings. For example, \cite{zhu2023xtab,hollmanntabpfn} propose a pretraining strategy for tabular Transformers that enables generalization to unseen columns, highlighting a promising path toward more flexible and scalable tabular learning frameworks.

\subsection{Graph Neural Networks} \label{sub:gnns}
Graph Neural Networks are powerful architectures that perform deep learning on graphs by jointly processing the connectivity information in a networks along with additional feature information at the node and/or edge level. There is a variety of different architectures that have been proposed, among which, the most popular are the message-passing GNNs \cite{kipf2016semi, hamilton2017inductive, gilmer2017neural, velickovic2017graph}. Standard message-passing GNNs are defined by the following recursive formula:
\begin{equation}\label{eq:GNNrec00}
    \mathbf{x}_v^{(l)} = g^{(l-1)}\left(\mathbf{x}_v^{(l-1)},f^{(l-1)}\left(\left\{\mathbf{x}_u^{(l-1)}:u\in\mathcal{N}\left(v\right)\right\}\right)\right),
\end{equation}
where, $\mathcal{N}(v)$ represents the 1-hop neighborhood of vertex $v$. The function $f^{(l)}$ aggregates information from the multiset of signals coming from neighboring vertices, while $g^{(l)}$ combines the signal of each vertex with the aggregated information from its neighbors. Common choices for $f^{(l)}$ and $g^{(l)}$ include the single- and multi-layer perceptron (MLP), the linear function, and the summation function. 
Foundational works in message-passing GNNs include Graph Convolutional Networks (GCNs) \cite{kipf2016semi}, Graph Isomorphism Networks (GINs) \cite{xu2018powerful}, and GraphSAGE \cite{hamilton2017inductive}.

The properties of GNNs have been extensively studied across multiple works. Notably, \cite{maron2018invariant} examines their permutation invariance and equivariance properties, while \cite{gama2020} analyzes their stability under perturbations. The transferability of GNNs has been explored in \cite{Ruiz2020,levie2021transferability}. Their expressive power, particularly in the context of graph isomorphism, has been studied in \cite{morris2019weisfeiler,xu2018powerful,kanatsoulis2024graph}. Furthermore, GNNs' ability to perform and generalize substructure counting is discussed in \cite{arvind2020weisfeiler,chen2020can,kanatsoulis2024counting}, while their capacity to solve graph bi-connectivity is examined in \cite{zhang2023rethinking}. 
\vspace{-0.4cm}
\subsection{Temporal Graph Neural Networks} \label{sub:tgnns}
Temporal Graph Neural Networks~(TGNNs) have emerged as promising architectures for modeling evolving graphs, also known as temporal graphs. Recently, TGNNs have achieved success in traffic forecasting~\cite{cini2023scalable}, link prediction~\cite{xu2020inductive}, behavior modeling~\cite{huang2023temporal}, node classification~\cite{rossi2020temporal} and financial analysis~\cite{shamsi2024graphpulse}. TGNNs model graph and time information, both of which are relevant to RDL tasks that predict future properties in a database. TGNNs are categorized based on their input data types: Continuous-Time Dynamic Graph (CTDG) and Discrete-Time Dynamic Graph (DTDG) methods.

\textbf{DTDG Methods.} Discrete-time approaches handle temporal graphs as sequences of static graph snapshots at regular intervals. DTDG methods often consists of two main modules, a recurrent module for modeling temporal dependencies such as RNN~\cite{rumelhart1985learning}, GRU~\cite{cho2014learning} and LSTM~\cite{hochreiter1997long} and a spatial module (usually a GNN) for capturing the graph topology at each snapshot.  
DTDG methods often focus on predicting future graph states or modeling temporal dependencies between consecutive snapshots. While simpler to implement, these methods may miss fine-grained temporal patterns between snapshots.
One example DTDG method is ROLAND, introduced by \citet{you2022roland}. The ROLAND framework enables the adaptation of static GNNs to evolving graph structures. At its core, ROLAND treats node embeddings from different GNN layers as hierarchical states, which are recurrently updated over time using recurrent models to incorporate new graph snapshots. This design allows seamless integration of state-of-the-art static GNN architectures while capturing temporal dynamics. 

\textbf{CTDG Methods.} Continuous-time approaches model temporal graphs as streams of events or interactions occurring at precise timestamps. CTDG methods often employ a GNN module for processing graph topology, a memory module for learning temporal dependencies and a temporal neighbor sampling scheme for retrieving past edges relevant to the current prediction. Example methods include TGAT~\cite{xu2020inductive}, which pioneered inductive representation learning using functional time encoding and temporal attention mechanisms. TGN~\cite{rossi2020temporal} introduced a versatile framework combining memory modules with graph operators for efficient processing of temporal event streams. CAWN~\cite{wang2020inductive} proposed using causal anonymous walks to automatically capture temporal network motifs. More recent advances include TCL~\cite{wang2021tcl} which uses a graph-topology-aware transformer architecture, and NAT~\cite{luo2022neighborhood} which employs efficient neighborhood representation through dictionary-type structures. 

\subsection{Graph Transformers} \label{sub:gts}
Transformers \cite{vaswani2017attention} were initially designed for natural language processing, where the self-attention mechanism enables context-aware sequence modeling. 
Motivated by their success, there have been significant efforts to develop powerful Transformer networks for graphs which can impact how we design RDL architectures.

\textbf{Transformers as GNNs.} Early attempts to adapt Transformers to graph-structured data include the development of Graph Attention Networks (GATs) \cite{velickovic2017graph} and Graph Transformers (GTs) \cite{dwivedi2021generalization} that bias the attention mechanism to a node’s local neighborhood. Formally, such trivial extension of Transformers on graphs can be derived as a special case of Eqn. \ref{eq:GNNrec00} where the message-passing aggregation function is instantiated with self-attention \cite{joshi2020transformers, bronstein2021geometric}. The GTs in \citet{dwivedi2021generalization} also extend the positional encodings (PEs) of Transformers \cite{vaswani2017attention} to arbitrary graph structures using Laplacian eigenvectors \cite{dwivedi2020benchmarking}, noting that the sinusoidal PEs in Transformers are a special case of Laplacian eigenvectors on a line graph (the underlying structure of sequences).

\textbf{Global attention.} Another line of works on GTs focus on applying the all-pair attention operation to graph nodes directly \cite{ying2021transformers, mialon2021graphit, kreuzer2021rethinking}, where nodes are often treated as tokens while edges (or adjacency matrices) provide structural biases in the form of absolute and relative positional encodings. Such methods capture global node interactions more effectively than traditional message-passing approaches, which is commonly limited to local neighborhood \cite{ying2021transformers}. However, they also give rise to significant computational challenges: the quadratic cost of all-pair attention on large, sparse graphs make Transformers expensive in both memory and runtime, motivating work on approximate or tailored attention schemes \cite{rampavsek2022recipe}.

\textbf{Modern GTs.} Recent efforts have therefore addressed two key obstacles: designing positional/structural encodings (PE/SE) that accommodate graph topology, and ensuring scalability to large graphs through a broader non-local attention coverage, such as GraphGPS~\cite{rampavsek2022recipe}. For the former, techniques such as relative distance embeddings, random walk features, Laplacian eigenvectors or learnable positional encoders~\cite{dwivedi2021generalization,ying2021transformers,kreuzer2021rethinking, dwivedi2022graph,canturk2023graph,limsign,huangstability,kanatsoulis2025learning} provide Transformers with positional and structural awareness in graph structured data. For scalability, hierarchical or clustering modules group nodes into coarser super-nodes \cite{zhang2022hierarchical,zhu2023hierarchical}, and sparse attention mechanisms limit the cost of dense global attention~\cite{rampavsek2022recipe,shirzad2023exphormer}. Meanwhile, strategies like neighborhood sampling and multi-hop tokenization~\cite{zhao2021gophormer,chen2022nagphormer,kong2023goat,dwivedi2023graph} offer practical ways to handle giant graphs, albeit sometimes restricting or approximating receptive fields. These hybrid local-global designs aim to preserve the global 
context advantages of Transformers while ensuring efficiency and reasonable memory footprints when scaling to large graph sizes.
\vspace{-0.2cm}
\subsection{Heterogenous Graph Architectures} \label{sub:hgnns}
Message-passing GNNs and Graph Attention Networks have been extended to handle heterogeneous graphs, which encompass complex structures where nodes and edges belong to multiple types. Unlike homogeneous GNNs that operate under the assumption of uniform node and edge types, heterogeneous GNNs \cite{schlichtkrull2018modeling,wang2019heterogeneous} incorporate specialized aggregation techniques, such as relation-aware message passing and metapath-based methods, to effectively capture the intricate relationships in real-world graphs. Building on this foundation, heterogeneous Graph Transformers \cite{hu2020heterogeneous} further enhance representation learning by employing attention mechanisms that dynamically adjust to different node and edge types and achieved strong performance across various applications, including recommendation systems and knowledge graph reasoning.

For example, RGCN extends standard GCNs to handle multi-relational and heterogeneous graphs by incorporating relation-specific transformations during message passing~\cite{schlichtkrull2018modeling}. Unlike homogeneous GCNs, which apply a single weight matrix to all edges, RGCN assigns distinct weight matrices to each edge type, preserving relational semantics in tasks like node classification and link prediction. To manage parameter complexity with numerous relations, RGCN employs basis decomposition, which shares weights across relations via linear combinations of basis matrices. This architecture enables RGCN to model heterogeneous graphs by aggregating neighbor features through relation-dependent pathways, capturing type-specific interactions. 

HAN~\cite{wang2019heterogeneous} is a framework for learning node representations in heterogeneous graphs by hierarchically attending to node and meta-path level semantics. HAN first projects nodes of different types into a shared feature space using type-specific transformation matrices, enabling cross-type feature integration. It then employs a two-tier attention mechanism: node-Level attention and semantic-level attention. Node-level attention aggregates features from meta-path-defined neighbors, dynamically weighting their importance via multi-head attention. Semantic-Level attention learns task-specific weights for distinct meta-paths  fusing their embeddings to capture diverse relational semantics. This dual attention enables HAN to handle heterogeneity while maintaining linear computational complexity relative to meta-path pairs, ensuring scalability.

In summary, traditional tabular models provide efficient baselines yet fail to fully exploit the relational structures present in multi-table settings. Graph neural networks overcome this limitation by directly incorporating the structural connectivity between entities from different tables. Extending GNNs to temporal and heterogeneous graph modeling further improves their modeling capabilities, capturing dynamic interactions and diverse entity types prevalent in relational databases. Integrating these methods with recent Transformer-based architectures offers a promising pathway towards unified, scalable models for Relational Deep Learning.

\section{Frontiers of RDL}
\label{sec:frontiers}

In this section, we first review state-of-the-art architectures specifically designed for RDL, including advanced graph neural networks (Section~\ref{sub:arch}) and large language models (Section~\ref{sub:llm}). We then discuss how recent developments in temporal graph learning-such as time encoding strategies (Section~\ref{sub:time}) and temporal graph methods (Section~\ref{sub:tgl})-can be leveraged to address RDL challenges.
\vspace{-0.2cm}
\subsection{Advanced RDL Architectures} 
\label{sub:arch}
\textbf{Graph Neural Networks:} Among the first implementations of RDL \cite{fey2023relational,robinson2025relbench}, a GNN based method encodes heterogeneous features from diverse table columns using PyTorch Frame \cite{hu2024pytorch}, which transforms raw row-level information into initial node embeddings. These embeddings are then input into a heterogeneous GraphSAGE model \cite{hamilton2017inductive, robinson2025relbench}, which updates them iteratively using a temporally-aware neighbor aggregation mechanism. In particular, a temporal-aware subgraph sampling is utilized to enforce causal constraints during training. For a given seed time, a subgraph is constructed by selecting only those neighboring nodes that existed prior to the seed time, thereby preventing information leakage. The GNN processes this sampled subgraph along with the node representations produced by PyTorch Frame, generating refined embeddings that are subsequently processed by task-specific prediction heads.

To leverage the unique structural properties of relational graphs, RelGNN \cite{chen2025relgnn} introduces a specialized graph attention mechanism to generate entity embeddings. The architecture is motivated by the observation that primary-foreign key relationships in relational graphs frequently give rise to two structural patterns: (i) bridge nodes, which connect exactly two foreign keys and form tripartite structures, and (ii) hub nodes, which link three or more foreign keys, creating star-shaped subgraphs. These nodes primarily act as intermediaries that facilitate interactions between their neighbors, and can lead to significant modeling inefficiencies and information loss. To address these challenges, RelGNN introduces atomic routes, which define structured sequences of node types that enable direct information exchange between the neighbors of bridge and hub nodes. By performing composite message-passing and graph attention over these atomic routes, RelGNN selectively aggregates relevant information, thus preserving critical relational structures.

ContextGNN \cite{yuan2024contextgnn} is introduced as a framework to improve recommendation tasks in relational databases by addressing key limitations in conventional models. First, two-tower models fail to capture contextual dependencies, leading to pair-agnostic representations. Second, while pair-wise representation models encode richer contextual information, they often suffer from scalability issues due to their quadratic complexity or restrictive constraints on candidate pairs. To overcome these, ContextGNN employs a hybrid approach that combines both strategies. It utilizes pair-wise representations to model relationships among familiar items within a user’s local subgraph while leveraging two-tower representations to facilitate recommendations for novel or less frequently encountered items. A final prediction network then integrates these representations, effectively balancing accuracy and scalability in ranking predictions.

TREeLGNN \cite{lachiover} combines the strengths of pretrained single-table models with static GNNs, in a time-then-graph manner, to improve performance and achieve significant inference speedup for RDL. In Particular, TREeLGNN leverages the power of tabular models to model the temporal dynamics for each table in the relational database. Then a static GNN is employed to capture complex relationships between different tables within the database.

\textbf{Transformers:}
Transformers have been successfully applied to single-table tabular data \cite{hollmanntabpfn,zhu2023xtab}, but extending them to relational databases with complex multi-table relationships presents significant challenges. To address this, \citet{pelevska2024transformers} introduced DBFormer which consists of two key components. First, it transforms each row of every table into meaningful unified embeddings using inter-table self-attention. Next, a cross-attention layer aggregates embeddings based on primary-foreign key relationships, allowing the model to effectively capture dependencies between related tables. This step is reminiscent of the operations performed in GATs \cite{velickovic2017graph}. \citet{dwivedi2025relational} proposed the Relational Graph Transformer (RelGT), a graph-transformer architecture tailored to relational databases. To encode heterogeneity, temporality, and topology efficiently, RelGT tokenizes each node into five elements-- features, type, hop distance, timestamp, and local structure. It then adopts a hybrid representation scheme, pairing local attention on sampled subgraphs with global attention to a small set of learnable centroids, thus unifying local and database-wide information.

An equally crucial aspect of transformers for effectively encoding data structure is PEs. While PEs specifically designed for relational databases have not been extensively explored, prior research \cite{kanatsoulis2025learning, bevilacquaholographic} has introduced PE frameworks that benefit RDL. Notably, \citet{kanatsoulis2025learning} identify four key properties that graph PEs should satisfy: stability, expressive power, scalability, and generality. However, existing eigenvector-based PE methods often struggle to meet all these criteria simultaneously. A learnable PE framework, PEARL \cite{kanatsoulis2025learning} is developed, leveraging a stable and expressive message-passing architecture, that approximates equivariant functions of eigenvectors with low complexity. PEARL’s properties, and empirical performance, make it a scalable and effective solution for relational data PEs. \citet{bevilacquaholographic} introduce holographic node representations, a scalable and adaptive PE approach that decouples expressiveness from task specificity. It comprises a task-agnostic expansion map that generates high-dimensional embeddings without node-permutation symmetries and a reduction map that selectively reinstates relevant symmetries. Such flexibility and scalability renders it a great candidate for relational database encoding.
\vspace{-0.2cm}
\subsection{Large Language Models for RDL} \label{sub:llm}
While GNN-based solutions have been prominently developed for RDL recently as reviewed in Section \ref{sub:arch}, large language models have recently shown promise for direct or in-context predictions. \citet{wydmuch2024tackling}  introduces a new baseline for RDL that leverages pre-trained LLMs to solve tasks on RelBench. The prediction problem is formulated as a text document comprising (i) a concise description of the relational database, (ii) a brief description of the prediction task, and (iii) a set of in-context examples. Each example includes the target label, timestamp, and a structured representation of entities, incorporating nested entities from related tables. These entities are serialized as JSON for input processing. For classification tasks, the approach employs metric-aware inference, while for regression tasks, a small multilayer perceptron head is trained using a limited subset of the training documents. 
 
\subsection{Time Encodings} \label{sub:time}

Relational data evolves over time where records in the tables naturally grow over time. Many tasks of interest involves the prediction of a query based on given timestamps (often in the future)~\cite{fey2023relational} thus designing an effective time encoding is crucial. The design of time encoding functions in temporal graph methods can be easily utilized in RDL approaches. Both learnable and non-learnable time encoding functions have been proposed in the literature.

\textbf{Learnable time encoding.}
Time2Vec~\cite{kazemi2019time2vec} is a common learnable time encoding used in state-of-the-art temporal graph models such as TGN~\cite{rossi2020temporal} and TGAT~\cite{xu2020inductive}. Its encoding function $\Phi : \mathcal{T} \rightarrow \mathbb{R}^d$ is a $d$ dimensional functional mapping from timestamps $t$ to $\mathbb{R}^d$.
\vspace{-5pt}
\begin{equation}
    t \rightarrow \Phi_d(t) := \sqrt{\frac{1}{d}} [\cos(\omega_1 t), \sin(\omega_1 t), \hdots, \cos(\omega_d t), \sin(\omega_d t)],
\end{equation}
where $\omega$ is a learnable frequency vector. The function aims to preserve relative time 
between timestamps, approximating a stationary temporal kernel $\mathcal{K}(t_i - t_j) = \langle \Phi_d(t_1), \Phi_d(t_2) \rangle$. The encoding derives from Bochner's theorem~\cite{loomis2013introduction}, which states that any stationary kernel can be represented as the Fourier transform of a non-negative measure, ensuring the theoretical validity of the encoding.

\textbf{Fixed time encoding.}
GraphMixer~\cite{congwe} proposes a novel non-learnable / fixed time encoding that has parameters dependent on the time range of the dataset while guaranteeing two important properties in time modeling. The time encoding is as follows:
\vspace{-3pt}
\begin{equation}
    t \rightarrow \cos(t\omega) \; \text{where} \; \omega = \{\alpha^{-(i-1)/\beta} \}_{i=1}^d
\end{equation}
The fixed time encoding has two beneficial properties. 
First, timestamps that are chronologically close to each other will map to similar embeddings. 
Second, timestamps that have high values are set to be a constant thus preserving recency bias.

\textbf{Understanding Time Granularity.} In RDL, a given predictive task has a fixed time granularity (\textit{e.g.}, daily, weekly).
However, real-world predictive tasks vary in time granularity and sometimes require prediction across multiple time granularities. For example, when forecasting the total sales of an item, predictions for the next day, week, and month might all be beneficial for decision making. Therefore, understanding the variation in model performance between time granularities is an interesting direction. Recently, \citet{huangutg2024} introduced the Unified Temporal Graph~(UTG) framework which formulated the discretization operation on timestamps and introduced the notation of discretization Level, that describes how coarse the time granularity is for a given dataset. Future research can investigate the effect of time granularity in RDL tasks.

\subsection{Temporal Graph Learning} \label{sub:tgl}

RDL requires the modeling of both the relation between entities and their dependencies over time. As such, we review the recent advancements in temporal graph research here.

\textbf{Heterogeneous TGNNs.} Relational databases can be modeled as a temporal heterogeneous graph~(HTG)~(see Section~\ref{sub:rel_graph}). Therefore, HTG methods can be modified directly for RDL.
Recently, \citet{li2023simplifying} proposed STHN, a novel architecture for temporal heterogeneous graphs. The architecture consists of a \emph{Heterogeneous Link Encoder} with type and time encoding components, which embed historical interactions to produce a temporal link representation. The process continues with \emph{Semantic Patches Fusion}, where sequential representations are divided into different patches treated as token inputs for the Encoder, and average mean pooling compresses these into a single node embedding for downstream tasks.

\textbf{Temporal Graph Transformers.} Temporal GTs naturally model both temporal and structural information, and demonstrate strong performance for various tasks. 
Therefore, adapting temporal graph transformers for RDL is a promising direction. \citet{yu2023towards} introduce DyGFormer, a transformer-based architecture for emporal graphs. DyGFormer has two key components: a \emph{neighbor co-occurrence encoding} scheme that explores correlations between source and destination nodes based on their historical sequences, and a \emph{patching technique} that divides sequences into multiple patches for efficient processing by the Transformer. Another recent work is HOT~\cite{besta2024hot}, designed to model higher-order (HO) graph structures with transformers. For efficiency, HOT employs a hierarchical attention mechanism using Block-Recurrent Transformers (BRT).

To conclude, advanced GNN architectures, transformer-based models, and improved temporal learning in graphs collectively demonstrate significant progress in capturing complex structure in RDBs for end-to-end representation learning, thereby further strengthening the foundation of the RDL as a research program.

\section{Discussion and Future Directions}
\label{sec:future}

While recent works in relational deep learning is promising, numerous research opportunities remain for unlocking new frontiers in RDL that can enable the wider adoption of AI in enterprises where large volume of relational data exist. Here, we highlight the two key directions that cover multiple aspects of future developments.

\subsection{Unified GNNs}
\label{sub:unified_gnns}

Building on the advances in Sections \ref{sec:methods} and \ref{sec:frontiers}, a key open question in RDL is how to unify different methodological insights-from traditional tabular models, to heterogeneous and temporal GNNs-and Transformers into a single end-to-end framework that can flexibly handle multi-table relational data. 

\textbf{Challenges.} As discussed, specialized solutions exist for distinct facets of the problem (e.g., bridging tabular and graph models~\cite{lachiover}, designing relational message passing~\cite{chen2025relgnn}, enforcing temporal dependencies \cite{fey2023relational, robinson2025relbench}, or incorporating learned time encodings \cite{rossi2020temporal}). However, unified architectures that jointly capture these diverse dimensions of RDL remain in early stages of development. 

Relational entity graphs can contain millions of nodes, each with potentially diverse feature modalities (e.g., numerical, text, or images) derived from different tables \cite{hu2024pytorch}. Effective unification requires a single message-passing pipeline that can (i) respect relational keys (i.e., primary--foreign key relationships) as edges, (ii) prevent time leakage through temporal subgraph sampling, and (iii) integrate the specialized column encoders needed for multi-modal attributes. Moreover, relational databases often exhibit star-like or hub-like connectivity, implying that learned messages must be routed carefully to avoid overwhelming bridge nodes~\cite{chen2025relgnn}.

\textbf{Advantages.}
By consolidating these elements, unified GNNs can simplify model selection by bridging multiple architectures that previously addressed only a subset of RDL challenges. They also improve transferability, making it easier to adapt a single model across heterogeneous databases with minimal configuration, and enable larger-scale training by providing a standardized framework amenable to distributed or streaming implementations on massive relational databases.
We thus foresee unified GNNs as a critical direction for advancing RDL. By systematically integrating multi-modal column encoders, relational message passing, and temporal consistency in a single, scalable architecture, future methods will unlock new industrial applications and research possibilities from the ever-growing stores of complex multi-table data.

\subsection{Foundation Models}
\label{sub:foundation_models}

Motivated by the insights of using LLMs for relational databases (Section \ref{sub:llm}), an exciting opportunity is the development of foundation models tailored to multi-table data. A first step in this direction is the Kumo Relational Foundation Model \cite{feykumorfm}, an in-context learner designed for multi-table relational graph settings. Just as large-scale pretrained transformers transformed NLP by learning universal language representations~\cite{achiam2023gpt, team2023gemini}, similar models can be envisioned for RDL, leveraging the structural homogeneity of primary--foreign key relationships across domains such as e-commerce, social media, and healthcare \cite{fey2023relational, zahradnik2023deep}. By encoding rich table schemas, time-evolving interactions, and heterogeneous attributes into a single representation space, these models would enable zero-shot or few-shot adaptation to new tasks with minimal retraining. Moreover, relational databases typically have recurring schemas or similar entity relationships, making them more amenable to shared embedding strategies than entirely arbitrary graphs.

Realizing such foundation models demands scalable pipelines for ingesting massive, multi-modal relational databases along with effective self-supervised tasks (e.g., masked column or link prediction) to capture essential structural and temporal signals. Integrations with textual columns, retrieval augmentation, or hybrid GNN-LLM frameworks~\cite{wydmuch2024tackling} could further enhance generalization. Successful foundation models for RDL would unify data from different verticals within a single pipeline, offering universal embeddings and robust out-of-the-box predictive capabilities. This has the potential to lower the barrier for industrial application, accelerate scientific discovery in multi-table datasets, and lay the groundwork for broader AI systems that natively understand relational structures.

\vspace{-0.1cm}
\begin{acks}
  Shenyang Huang was supported by Natural Sciences and Engineering Research Council of Canada (NSERC) Postgraduate Scholarship Doctoral (PGS D) Award and Fonds de recherche du Québec - Nature et Technologies (FRQNT) Doctoral Award. The authors gratefully acknowledge the support of NSF under Nos. OAC-1835598 (CINES), CCF-1918940 (Expeditions), DMS-2327709 (IHBEM), IIS-2403318 (III); Stanford Data Applications Initiative, Wu Tsai Neurosciences Institute, Stanford Institute for Human-Centered AI, Chan Zuckerberg Initiative, Amazon, Genentech, Hitachi, and SAP. The content is solely the responsibility of the authors and does not necessarily represent the official views of the funding entities.
\end{acks}

\bibliographystyle{ACM-Reference-Format}
\bibliography{ref}


\begin{thebibliography}{105}


\ifx \showCODEN    \undefined \def \showCODEN     #1{\unskip}     \fi
\ifx \showISBNx    \undefined \def \showISBNx     #1{\unskip}     \fi
\ifx \showISBNxiii \undefined \def \showISBNxiii  #1{\unskip}     \fi
\ifx \showISSN     \undefined \def \showISSN      #1{\unskip}     \fi
\ifx \showLCCN     \undefined \def \showLCCN      #1{\unskip}     \fi
\ifx \shownote     \undefined \def \shownote      #1{#1}          \fi
\ifx \showarticletitle \undefined \def \showarticletitle #1{#1}   \fi
\ifx \showURL      \undefined \def \showURL       {\relax}        \fi
\providecommand\bibfield[2]{#2}
\providecommand\bibinfo[2]{#2}
\providecommand\natexlab[1]{#1}
\providecommand\showeprint[2][]{arXiv:#2}

\bibitem[Achiam et~al\mbox{.}(2023)]%
        {achiam2023gpt}
\bibfield{author}{\bibinfo{person}{Josh Achiam}, \bibinfo{person}{Steven Adler}, \bibinfo{person}{Sandhini Agarwal}, \bibinfo{person}{Lama Ahmad}, \bibinfo{person}{Ilge Akkaya}, \bibinfo{person}{Florencia~Leoni Aleman}, \bibinfo{person}{Diogo Almeida}, \bibinfo{person}{Janko Altenschmidt}, \bibinfo{person}{Sam Altman}, \bibinfo{person}{Shyamal Anadkat}, {et~al\mbox{.}}} \bibinfo{year}{2023}\natexlab{}.
\newblock \showarticletitle{Gpt-4 technical report}.
\newblock \bibinfo{journal}{\emph{arXiv preprint arXiv:2303.08774}} (\bibinfo{year}{2023}).
\newblock


\bibitem[Anthropic(2024)]%
        {anthropic2024claude}
\bibfield{author}{\bibinfo{person}{AI Anthropic}.} \bibinfo{year}{2024}\natexlab{}.
\newblock \showarticletitle{The claude 3 model family: Opus, sonnet, haiku}.
\newblock \bibinfo{journal}{\emph{Claude-3 Model Card}}  \bibinfo{volume}{1} (\bibinfo{year}{2024}), \bibinfo{pages}{1}.
\newblock


\bibitem[Arik and Pfister(2021)]%
        {arik2021tabnet}
\bibfield{author}{\bibinfo{person}{Sercan~{\"O} Arik} {and} \bibinfo{person}{Tomas Pfister}.} \bibinfo{year}{2021}\natexlab{}.
\newblock \showarticletitle{Tabnet: Attentive interpretable tabular learning}. In \bibinfo{booktitle}{\emph{Proceedings of the AAAI conference on artificial intelligence}}, Vol.~\bibinfo{volume}{35}. \bibinfo{pages}{6679--6687}.
\newblock


\bibitem[Arvind et~al\mbox{.}(2020)]%
        {arvind2020weisfeiler}
\bibfield{author}{\bibinfo{person}{Vikraman Arvind}, \bibinfo{person}{Frank Fuhlbr{\"u}ck}, \bibinfo{person}{Johannes K{\"o}bler}, {and} \bibinfo{person}{Oleg Verbitsky}.} \bibinfo{year}{2020}\natexlab{}.
\newblock \showarticletitle{On Weisfeiler-Leman invariance: Subgraph counts and related graph properties}.
\newblock \bibinfo{journal}{\emph{J. Comput. System Sci.}}  \bibinfo{volume}{113} (\bibinfo{year}{2020}), \bibinfo{pages}{42--59}.
\newblock


\bibitem[Bai et~al\mbox{.}(2023)]%
        {bai2023qwen}
\bibfield{author}{\bibinfo{person}{Jinze Bai}, \bibinfo{person}{Shuai Bai}, \bibinfo{person}{Yunfei Chu}, \bibinfo{person}{Zeyu Cui}, \bibinfo{person}{Kai Dang}, \bibinfo{person}{Xiaodong Deng}, \bibinfo{person}{Yang Fan}, \bibinfo{person}{Wenbin Ge}, \bibinfo{person}{Yu Han}, \bibinfo{person}{Fei Huang}, {et~al\mbox{.}}} \bibinfo{year}{2023}\natexlab{}.
\newblock \showarticletitle{Qwen technical report}.
\newblock \bibinfo{journal}{\emph{arXiv preprint arXiv:2309.16609}} (\bibinfo{year}{2023}).
\newblock


\bibitem[Besta et~al\mbox{.}(2024)]%
        {besta2024hot}
\bibfield{author}{\bibinfo{person}{Maciej Besta}, \bibinfo{person}{Afonso~Claudino Catarino}, \bibinfo{person}{Lukas Gianinazzi}, \bibinfo{person}{Nils Blach}, \bibinfo{person}{Piotr Nyczyk}, \bibinfo{person}{Hubert Niewiadomski}, {and} \bibinfo{person}{Torsten Hoefler}.} \bibinfo{year}{2024}\natexlab{}.
\newblock \showarticletitle{Hot: Higher-order dynamic graph representation learning with efficient transformers}. In \bibinfo{booktitle}{\emph{Learning on Graphs Conference}}. PMLR, \bibinfo{pages}{15--1}.
\newblock


\bibitem[Bevilacqua et~al\mbox{.}({[n.\,d.]})]%
        {bevilacquaholographic}
\bibfield{author}{\bibinfo{person}{Beatrice Bevilacqua}, \bibinfo{person}{Joshua Robinson}, \bibinfo{person}{Jure Leskovec}, {and} \bibinfo{person}{Bruno Ribeiro}.} \bibinfo{year}{[n.\,d.]}\natexlab{}.
\newblock \showarticletitle{Holographic Node Representations: Pre-training Task-Agnostic Node Embeddings}. In \bibinfo{booktitle}{\emph{The Thirteenth International Conference on Learning Representations}}.
\newblock


\bibitem[Bi et~al\mbox{.}(2024)]%
        {bi2024deepseek}
\bibfield{author}{\bibinfo{person}{Xiao Bi}, \bibinfo{person}{Deli Chen}, \bibinfo{person}{Guanting Chen}, \bibinfo{person}{Shanhuang Chen}, \bibinfo{person}{Damai Dai}, \bibinfo{person}{Chengqi Deng}, \bibinfo{person}{Honghui Ding}, \bibinfo{person}{Kai Dong}, \bibinfo{person}{Qiushi Du}, \bibinfo{person}{Zhe Fu}, {et~al\mbox{.}}} \bibinfo{year}{2024}\natexlab{}.
\newblock \showarticletitle{Deepseek llm: Scaling open-source language models with longtermism}.
\newblock \bibinfo{journal}{\emph{arXiv preprint arXiv:2401.02954}} (\bibinfo{year}{2024}).
\newblock


\bibitem[Bronstein et~al\mbox{.}(2021)]%
        {bronstein2021geometric}
\bibfield{author}{\bibinfo{person}{Michael~M Bronstein}, \bibinfo{person}{Joan Bruna}, \bibinfo{person}{Taco Cohen}, {and} \bibinfo{person}{Petar Veli{\v{c}}kovi{\'c}}.} \bibinfo{year}{2021}\natexlab{}.
\newblock \showarticletitle{Geometric deep learning: Grids, groups, graphs, geodesics, and gauges}.
\newblock \bibinfo{journal}{\emph{arXiv preprint arXiv:2104.13478}} (\bibinfo{year}{2021}).
\newblock


\bibitem[Cant{\"u}rk et~al\mbox{.}(2023)]%
        {canturk2023graph}
\bibfield{author}{\bibinfo{person}{Semih Cant{\"u}rk}, \bibinfo{person}{Renming Liu}, \bibinfo{person}{Olivier Lapointe-Gagn{\'e}}, \bibinfo{person}{Vincent L{\'e}tourneau}, \bibinfo{person}{Guy Wolf}, \bibinfo{person}{Dominique Beaini}, {and} \bibinfo{person}{Ladislav Ramp{\'a}{\v{s}}ek}.} \bibinfo{year}{2023}\natexlab{}.
\newblock \showarticletitle{Graph positional and structural encoder}.
\newblock \bibinfo{journal}{\emph{arXiv preprint arXiv:2307.07107}} (\bibinfo{year}{2023}).
\newblock


\bibitem[Chen et~al\mbox{.}(2022)]%
        {chen2022nagphormer}
\bibfield{author}{\bibinfo{person}{Jinsong Chen}, \bibinfo{person}{Kaiyuan Gao}, \bibinfo{person}{Gaichao Li}, {and} \bibinfo{person}{Kun He}.} \bibinfo{year}{2022}\natexlab{}.
\newblock \showarticletitle{NAGphormer: A tokenized graph transformer for node classification in large graphs}. In \bibinfo{booktitle}{\emph{The Eleventh International Conference on Learning Representations}}.
\newblock


\bibitem[Chen et~al\mbox{.}(2023)]%
        {chen2023trompt}
\bibfield{author}{\bibinfo{person}{Kuan-Yu Chen}, \bibinfo{person}{Ping-Han Chiang}, \bibinfo{person}{Hsin-Rung Chou}, \bibinfo{person}{Ting-Wei Chen}, {and} \bibinfo{person}{Tien-Hao Chang}.} \bibinfo{year}{2023}\natexlab{}.
\newblock \showarticletitle{Trompt: Towards a Better Deep Neural Network for Tabular Data}.
\newblock


\bibitem[Chen and Guestrin(2016)]%
        {chen2016xgboost}
\bibfield{author}{\bibinfo{person}{Tianqi Chen} {and} \bibinfo{person}{Carlos Guestrin}.} \bibinfo{year}{2016}\natexlab{}.
\newblock \showarticletitle{Xgboost: A scalable tree boosting system}. In \bibinfo{booktitle}{\emph{Proceedings of the 22nd acm sigkdd international conference on knowledge discovery and data mining}}. \bibinfo{pages}{785--794}.
\newblock


\bibitem[Chen et~al\mbox{.}(2025)]%
        {chen2025relgnn}
\bibfield{author}{\bibinfo{person}{Tianlang Chen}, \bibinfo{person}{Charilaos Kanatsoulis}, {and} \bibinfo{person}{Jure Leskovec}.} \bibinfo{year}{2025}\natexlab{}.
\newblock \showarticletitle{RelGNN: Composite Message Passing for Relational Deep Learning}.
\newblock \bibinfo{journal}{\emph{arXiv preprint arXiv:2502.06784}} (\bibinfo{year}{2025}).
\newblock


\bibitem[Chen et~al\mbox{.}(2020)]%
        {chen2020can}
\bibfield{author}{\bibinfo{person}{Zhengdao Chen}, \bibinfo{person}{Lei Chen}, \bibinfo{person}{Soledad Villar}, {and} \bibinfo{person}{Joan Bruna}.} \bibinfo{year}{2020}\natexlab{}.
\newblock \showarticletitle{Can graph neural networks count substructures?}
\newblock \bibinfo{journal}{\emph{Advances in neural information processing systems}}  \bibinfo{volume}{33} (\bibinfo{year}{2020}), \bibinfo{pages}{10383--10395}.
\newblock


\bibitem[Cho et~al\mbox{.}(2014)]%
        {cho2014learning}
\bibfield{author}{\bibinfo{person}{Kyunghyun Cho}, \bibinfo{person}{Bart van Merri{\"e}nboer}, \bibinfo{person}{{\c{C}}a{\u{g}}lar Gu̇l{\c{c}}ehre}, \bibinfo{person}{Dzmitry Bahdanau}, \bibinfo{person}{Fethi Bougares}, \bibinfo{person}{Holger Schwenk}, {and} \bibinfo{person}{Yoshua Bengio}.} \bibinfo{year}{2014}\natexlab{}.
\newblock \showarticletitle{Learning Phrase Representations using RNN Encoder--Decoder for Statistical Machine Translation}. In \bibinfo{booktitle}{\emph{Proceedings of the 2014 Conference on Empirical Methods in Natural Language Processing (EMNLP)}}. \bibinfo{pages}{1724--1734}.
\newblock


\bibitem[Cini et~al\mbox{.}(2023)]%
        {cini2023scalable}
\bibfield{author}{\bibinfo{person}{Andrea Cini}, \bibinfo{person}{Ivan Marisca}, \bibinfo{person}{Filippo~Maria Bianchi}, {and} \bibinfo{person}{Cesare Alippi}.} \bibinfo{year}{2023}\natexlab{}.
\newblock \showarticletitle{Scalable spatiotemporal graph neural networks}. In \bibinfo{booktitle}{\emph{Proceedings of the AAAI conference on artificial intelligence}}, Vol.~\bibinfo{volume}{37}. \bibinfo{pages}{7218--7226}.
\newblock


\bibitem[Cong et~al\mbox{.}({[n.\,d.]})]%
        {congwe}
\bibfield{author}{\bibinfo{person}{Weilin Cong}, \bibinfo{person}{Si Zhang}, \bibinfo{person}{Jian Kang}, \bibinfo{person}{Baichuan Yuan}, \bibinfo{person}{Hao Wu}, \bibinfo{person}{Xin Zhou}, \bibinfo{person}{Hanghang Tong}, {and} \bibinfo{person}{Mehrdad Mahdavi}.} \bibinfo{year}{[n.\,d.]}\natexlab{}.
\newblock \showarticletitle{Do We Really Need Complicated Model Architectures For Temporal Networks?}. In \bibinfo{booktitle}{\emph{The Eleventh International Conference on Learning Representations}}.
\newblock


\bibitem[Derrow-Pinion et~al\mbox{.}(2021)]%
        {derrow2021eta}
\bibfield{author}{\bibinfo{person}{Austin Derrow-Pinion}, \bibinfo{person}{Jennifer She}, \bibinfo{person}{David Wong}, \bibinfo{person}{Oliver Lange}, \bibinfo{person}{Todd Hester}, \bibinfo{person}{Luis Perez}, \bibinfo{person}{Marc Nunkesser}, \bibinfo{person}{Seongjae Lee}, \bibinfo{person}{Xueying Guo}, \bibinfo{person}{Brett Wiltshire}, {et~al\mbox{.}}} \bibinfo{year}{2021}\natexlab{}.
\newblock \showarticletitle{Eta prediction with graph neural networks in google maps}. In \bibinfo{booktitle}{\emph{Proceedings of the 30th ACM international conference on information \& knowledge management}}. \bibinfo{pages}{3767--3776}.
\newblock


\bibitem[Dwivedi and Bresson(2021)]%
        {dwivedi2021generalization}
\bibfield{author}{\bibinfo{person}{Vijay~Prakash Dwivedi} {and} \bibinfo{person}{Xavier Bresson}.} \bibinfo{year}{2021}\natexlab{}.
\newblock \showarticletitle{A Generalization of Transformer Networks to Graphs}.
\newblock \bibinfo{journal}{\emph{AAAI Workshop on Deep Learning on Graphs: Methods and Applications}} (\bibinfo{year}{2021}).
\newblock


\bibitem[Dwivedi et~al\mbox{.}(2025)]%
        {dwivedi2025relational}
\bibfield{author}{\bibinfo{person}{Vijay~Prakash Dwivedi}, \bibinfo{person}{Sri Jaladi}, \bibinfo{person}{Yangyi Shen}, \bibinfo{person}{Federico L{\'o}pez}, \bibinfo{person}{Charilaos~I Kanatsoulis}, \bibinfo{person}{Rishi Puri}, \bibinfo{person}{Matthias Fey}, {and} \bibinfo{person}{Jure Leskovec}.} \bibinfo{year}{2025}\natexlab{}.
\newblock \showarticletitle{Relational Graph Transformer}.
\newblock \bibinfo{journal}{\emph{arXiv preprint arXiv:2505.10960}} (\bibinfo{year}{2025}).
\newblock


\bibitem[Dwivedi et~al\mbox{.}(2020)]%
        {dwivedi2020benchmarking}
\bibfield{author}{\bibinfo{person}{Vijay~Prakash Dwivedi}, \bibinfo{person}{Chaitanya~K Joshi}, \bibinfo{person}{Thomas Laurent}, \bibinfo{person}{Yoshua Bengio}, {and} \bibinfo{person}{Xavier Bresson}.} \bibinfo{year}{2020}\natexlab{}.
\newblock \showarticletitle{Benchmarking graph neural networks}.
\newblock \bibinfo{journal}{\emph{arXiv:2003.00982}} (\bibinfo{year}{2020}).
\newblock


\bibitem[Dwivedi et~al\mbox{.}(2023)]%
        {dwivedi2023graph}
\bibfield{author}{\bibinfo{person}{Vijay~Prakash Dwivedi}, \bibinfo{person}{Yozen Liu}, \bibinfo{person}{Anh~Tuan Luu}, \bibinfo{person}{Xavier Bresson}, \bibinfo{person}{Neil Shah}, {and} \bibinfo{person}{Tong Zhao}.} \bibinfo{year}{2023}\natexlab{}.
\newblock \showarticletitle{Graph transformers for large graphs}.
\newblock \bibinfo{journal}{\emph{arXiv preprint arXiv:2312.11109}} (\bibinfo{year}{2023}).
\newblock


\bibitem[Dwivedi et~al\mbox{.}(2022)]%
        {dwivedi2022graph}
\bibfield{author}{\bibinfo{person}{Vijay~Prakash Dwivedi}, \bibinfo{person}{Anh~Tuan Luu}, \bibinfo{person}{Thomas Laurent}, \bibinfo{person}{Yoshua Bengio}, {and} \bibinfo{person}{Xavier Bresson}.} \bibinfo{year}{2022}\natexlab{}.
\newblock \showarticletitle{Graph Neural Networks with Learnable Structural and Positional Representations}. In \bibinfo{booktitle}{\emph{International Conference on Learning Representations}}.
\newblock


\bibitem[Ferrini et~al\mbox{.}(2024)]%
        {ferrini2024self}
\bibfield{author}{\bibinfo{person}{Francesco Ferrini}, \bibinfo{person}{Antonio Longa}, \bibinfo{person}{Andrea Passerini}, {and} \bibinfo{person}{Manfred Jaeger}.} \bibinfo{year}{2024}\natexlab{}.
\newblock \showarticletitle{A Self-Explainable Heterogeneous GNN for Relational Deep Learning}.
\newblock \bibinfo{journal}{\emph{arXiv preprint arXiv:2412.00521}} (\bibinfo{year}{2024}).
\newblock


\bibitem[Fey et~al\mbox{.}(2024)]%
        {fey2023relational}
\bibfield{author}{\bibinfo{person}{Matthias Fey}, \bibinfo{person}{Weihua Hu}, \bibinfo{person}{Kexin Huang}, \bibinfo{person}{Jan~Eric Lenssen}, \bibinfo{person}{Rishabh Ranjan}, \bibinfo{person}{Joshua Robinson}, \bibinfo{person}{Rex Ying}, \bibinfo{person}{Jiaxuan You}, {and} \bibinfo{person}{Jure Leskovec}.} \bibinfo{year}{2024}\natexlab{}.
\newblock \showarticletitle{Position: Relational deep learning-graph representation learning on relational databases}. In \bibinfo{booktitle}{\emph{Forty-first International Conference on Machine Learning}}.
\newblock


\bibitem[Fey et~al\mbox{.}({[n.\,d.]})]%
        {feykumorfm}
\bibfield{author}{\bibinfo{person}{Matthias Fey}, \bibinfo{person}{Vid Kocijan}, \bibinfo{person}{Federico Lopez}, \bibinfo{person}{Jan~Eric Lenssen}, {and} \bibinfo{person}{Jure Leskovec}.} \bibinfo{year}{[n.\,d.]}\natexlab{}.
\newblock \showarticletitle{KumoRFM: A Foundation Model for In-Context Learning on Relational Data}.
\newblock  (\bibinfo{year}{[n.\,d.]}).
\newblock


\bibitem[Galkin et~al\mbox{.}(2023)]%
        {galkin2023towards}
\bibfield{author}{\bibinfo{person}{Mikhail Galkin}, \bibinfo{person}{Xinyu Yuan}, \bibinfo{person}{Hesham Mostafa}, \bibinfo{person}{Jian Tang}, {and} \bibinfo{person}{Zhaocheng Zhu}.} \bibinfo{year}{2023}\natexlab{}.
\newblock \showarticletitle{Towards foundation models for knowledge graph reasoning}.
\newblock \bibinfo{journal}{\emph{arXiv preprint arXiv:2310.04562}} (\bibinfo{year}{2023}).
\newblock


\bibitem[Gallagher and Eliassi-Rad(2008)]%
        {gallagher2008leveraging}
\bibfield{author}{\bibinfo{person}{Brian Gallagher} {and} \bibinfo{person}{Tina Eliassi-Rad}.} \bibinfo{year}{2008}\natexlab{}.
\newblock \showarticletitle{Leveraging label-independent features for classification in sparsely labeled networks: An empirical study}. In \bibinfo{booktitle}{\emph{International Workshop on Social Network Mining and Analysis}}. Springer, \bibinfo{pages}{1--19}.
\newblock


\bibitem[Gama et~al\mbox{.}(2020)]%
        {gama2020}
\bibfield{author}{\bibinfo{person}{Fernando Gama}, \bibinfo{person}{Joan Bruna}, {and} \bibinfo{person}{Alejandro Ribeiro}.} \bibinfo{year}{2020}\natexlab{}.
\newblock \showarticletitle{Stability Properties of Graph Neural Networks}.
\newblock \bibinfo{journal}{\emph{IEEE Transactions on Signal Processing}}  \bibinfo{volume}{68} (\bibinfo{year}{2020}), \bibinfo{pages}{5680--5695}.
\newblock


\bibitem[Gilmer et~al\mbox{.}(2017)]%
        {gilmer2017neural}
\bibfield{author}{\bibinfo{person}{Justin Gilmer}, \bibinfo{person}{Samuel~S Schoenholz}, \bibinfo{person}{Patrick~F Riley}, \bibinfo{person}{Oriol Vinyals}, {and} \bibinfo{person}{George~E Dahl}.} \bibinfo{year}{2017}\natexlab{}.
\newblock \showarticletitle{Neural message passing for quantum chemistry}. In \bibinfo{booktitle}{\emph{International conference on machine learning}}. PMLR, \bibinfo{pages}{1263--1272}.
\newblock


\bibitem[Gorishniy et~al\mbox{.}(2022)]%
        {gorishniy2022embeddings}
\bibfield{author}{\bibinfo{person}{Yury Gorishniy}, \bibinfo{person}{Ivan Rubachev}, {and} \bibinfo{person}{Artem Babenko}.} \bibinfo{year}{2022}\natexlab{}.
\newblock \showarticletitle{On embeddings for numerical features in tabular deep learning}.
\newblock \bibinfo{journal}{\emph{Advances in Neural Information Processing Systems}}  \bibinfo{volume}{35} (\bibinfo{year}{2022}), \bibinfo{pages}{24991--25004}.
\newblock


\bibitem[Gorishniy et~al\mbox{.}(2021)]%
        {gorishniy2021revisiting}
\bibfield{author}{\bibinfo{person}{Yury Gorishniy}, \bibinfo{person}{Ivan Rubachev}, \bibinfo{person}{Valentin Khrulkov}, {and} \bibinfo{person}{Artem Babenko}.} \bibinfo{year}{2021}\natexlab{}.
\newblock \showarticletitle{Revisiting deep learning models for tabular data}, Vol.~\bibinfo{volume}{34}. \bibinfo{pages}{18932--18943}.
\newblock


\bibitem[Grattafiori et~al\mbox{.}(2024)]%
        {grattafiori2024llama}
\bibfield{author}{\bibinfo{person}{Aaron Grattafiori}, \bibinfo{person}{Abhimanyu Dubey}, \bibinfo{person}{Abhinav Jauhri}, \bibinfo{person}{Abhinav Pandey}, \bibinfo{person}{Abhishek Kadian}, \bibinfo{person}{Ahmad Al-Dahle}, \bibinfo{person}{Aiesha Letman}, \bibinfo{person}{Akhil Mathur}, \bibinfo{person}{Alan Schelten}, \bibinfo{person}{Alex Vaughan}, {et~al\mbox{.}}} \bibinfo{year}{2024}\natexlab{}.
\newblock \showarticletitle{The llama 3 herd of models}.
\newblock \bibinfo{journal}{\emph{arXiv preprint arXiv:2407.21783}} (\bibinfo{year}{2024}).
\newblock


\bibitem[Grover and Leskovec(2016)]%
        {grover2016node2vec}
\bibfield{author}{\bibinfo{person}{Aditya Grover} {and} \bibinfo{person}{Jure Leskovec}.} \bibinfo{year}{2016}\natexlab{}.
\newblock \showarticletitle{node2vec: Scalable feature learning for networks}. In \bibinfo{booktitle}{\emph{Proceedings of the 22nd ACM SIGKDD international conference on Knowledge discovery and data mining}}. \bibinfo{pages}{855--864}.
\newblock


\bibitem[Hamilton et~al\mbox{.}(2017)]%
        {hamilton2017inductive}
\bibfield{author}{\bibinfo{person}{Will Hamilton}, \bibinfo{person}{Zhitao Ying}, {and} \bibinfo{person}{Jure Leskovec}.} \bibinfo{year}{2017}\natexlab{}.
\newblock \showarticletitle{Inductive representation learning on large graphs}.
\newblock \bibinfo{journal}{\emph{Advances in neural information processing systems}}  \bibinfo{volume}{30} (\bibinfo{year}{2017}).
\newblock


\bibitem[Henderson et~al\mbox{.}(2012)]%
        {henderson2012rolx}
\bibfield{author}{\bibinfo{person}{Keith Henderson}, \bibinfo{person}{Brian Gallagher}, \bibinfo{person}{Tina Eliassi-Rad}, \bibinfo{person}{Hanghang Tong}, \bibinfo{person}{Sugato Basu}, \bibinfo{person}{Leman Akoglu}, \bibinfo{person}{Danai Koutra}, \bibinfo{person}{Christos Faloutsos}, {and} \bibinfo{person}{Lei Li}.} \bibinfo{year}{2012}\natexlab{}.
\newblock \showarticletitle{Rolx: structural role extraction \& mining in large graphs}. In \bibinfo{booktitle}{\emph{Proceedings of the 18th ACM SIGKDD international conference on Knowledge discovery and data mining}}. \bibinfo{pages}{1231--1239}.
\newblock


\bibitem[Hochreiter and Schmidhuber(1997)]%
        {hochreiter1997long}
\bibfield{author}{\bibinfo{person}{Sepp Hochreiter} {and} \bibinfo{person}{J{\"u}rgen Schmidhuber}.} \bibinfo{year}{1997}\natexlab{}.
\newblock \showarticletitle{Long short-term memory}.
\newblock \bibinfo{journal}{\emph{Neural computation}} \bibinfo{volume}{9}, \bibinfo{number}{8} (\bibinfo{year}{1997}), \bibinfo{pages}{1735--1780}.
\newblock


\bibitem[Hollmann et~al\mbox{.}({[n.\,d.]})]%
        {hollmanntabpfn}
\bibfield{author}{\bibinfo{person}{Noah Hollmann}, \bibinfo{person}{Samuel M{\"u}ller}, \bibinfo{person}{Katharina Eggensperger}, {and} \bibinfo{person}{Frank Hutter}.} \bibinfo{year}{[n.\,d.]}\natexlab{}.
\newblock \showarticletitle{TabPFN: A Transformer That Solves Small Tabular Classification Problems in a Second}. In \bibinfo{booktitle}{\emph{The Eleventh International Conference on Learning Representations}}.
\newblock


\bibitem[Hu et~al\mbox{.}(2020b)]%
        {hu2020open}
\bibfield{author}{\bibinfo{person}{Weihua Hu}, \bibinfo{person}{Matthias Fey}, \bibinfo{person}{Marinka Zitnik}, \bibinfo{person}{Yuxiao Dong}, \bibinfo{person}{Hongyu Ren}, \bibinfo{person}{Bowen Liu}, \bibinfo{person}{Michele Catasta}, {and} \bibinfo{person}{Jure Leskovec}.} \bibinfo{year}{2020}\natexlab{b}.
\newblock \showarticletitle{Open graph benchmark: Datasets for machine learning on graphs}.
\newblock \bibinfo{journal}{\emph{Advances in neural information processing systems}}  \bibinfo{volume}{33} (\bibinfo{year}{2020}), \bibinfo{pages}{22118--22133}.
\newblock


\bibitem[Hu et~al\mbox{.}(2024)]%
        {hu2024pytorch}
\bibfield{author}{\bibinfo{person}{Weihua Hu}, \bibinfo{person}{Yiwen Yuan}, \bibinfo{person}{Zecheng Zhang}, \bibinfo{person}{Akihiro Nitta}, \bibinfo{person}{Kaidi Cao}, \bibinfo{person}{Vid Kocijan}, \bibinfo{person}{Jinu Sunil}, \bibinfo{person}{Jure Leskovec}, {and} \bibinfo{person}{Matthias Fey}.} \bibinfo{year}{2024}\natexlab{}.
\newblock \showarticletitle{Pytorch frame: A modular framework for multi-modal tabular learning}.
\newblock \bibinfo{journal}{\emph{arXiv preprint arXiv:2404.00776}} (\bibinfo{year}{2024}).
\newblock


\bibitem[Hu et~al\mbox{.}(2020a)]%
        {hu2020heterogeneous}
\bibfield{author}{\bibinfo{person}{Ziniu Hu}, \bibinfo{person}{Yuxiao Dong}, \bibinfo{person}{Kuansan Wang}, {and} \bibinfo{person}{Yizhou Sun}.} \bibinfo{year}{2020}\natexlab{a}.
\newblock \showarticletitle{Heterogeneous graph transformer}. In \bibinfo{booktitle}{\emph{Proceedings of the web conference 2020}}. \bibinfo{pages}{2704--2710}.
\newblock


\bibitem[Huang et~al\mbox{.}(2023)]%
        {huang2023temporal}
\bibfield{author}{\bibinfo{person}{Shenyang Huang}, \bibinfo{person}{Farimah Poursafaei}, \bibinfo{person}{Jacob Danovitch}, \bibinfo{person}{Matthias Fey}, \bibinfo{person}{Weihua Hu}, \bibinfo{person}{Emanuele Rossi}, \bibinfo{person}{Jure Leskovec}, \bibinfo{person}{Michael Bronstein}, \bibinfo{person}{Guillaume Rabusseau}, {and} \bibinfo{person}{Reihaneh Rabbany}.} \bibinfo{year}{2023}\natexlab{}.
\newblock \showarticletitle{Temporal graph benchmark for machine learning on temporal graphs}.
\newblock \bibinfo{journal}{\emph{Advances in Neural Information Processing Systems}}  \bibinfo{volume}{36} (\bibinfo{year}{2023}), \bibinfo{pages}{2056--2073}.
\newblock


\bibitem[Huang et~al\mbox{.}({[n.\,d.]})]%
        {huangutg2024}
\bibfield{author}{\bibinfo{person}{Shenyang Huang}, \bibinfo{person}{Farimah Poursafaei}, \bibinfo{person}{Reihaneh Rabbany}, \bibinfo{person}{Guillaume Rabusseau}, {and} \bibinfo{person}{Emanuele Rossi}.} \bibinfo{year}{[n.\,d.]}\natexlab{}.
\newblock \showarticletitle{UTG: Towards a Unified View of Snapshot and Event Based Models for Temporal Graphs}. In \bibinfo{booktitle}{\emph{The Third Learning on Graphs Conference}}.
\newblock


\bibitem[Huang et~al\mbox{.}(2020)]%
        {huang2020tabtransformer}
\bibfield{author}{\bibinfo{person}{Xin Huang}, \bibinfo{person}{Ashish Khetan}, \bibinfo{person}{Milan Cvitkovic}, {and} \bibinfo{person}{Zohar Karnin}.} \bibinfo{year}{2020}\natexlab{}.
\newblock \showarticletitle{Tabtransformer: Tabular data modeling using contextual embeddings}.
\newblock \bibinfo{journal}{\emph{arXiv preprint arXiv:2012.06678}} (\bibinfo{year}{2020}).
\newblock


\bibitem[Huang et~al\mbox{.}(2024)]%
        {huangstability}
\bibfield{author}{\bibinfo{person}{Yinan Huang}, \bibinfo{person}{William Lu}, \bibinfo{person}{Joshua Robinson}, \bibinfo{person}{Yu Yang}, \bibinfo{person}{Muhan Zhang}, \bibinfo{person}{Stefanie Jegelka}, {and} \bibinfo{person}{Pan Li}.} \bibinfo{year}{2024}\natexlab{}.
\newblock \showarticletitle{On the Stability of Expressive Positional Encodings for Graphs}. In \bibinfo{booktitle}{\emph{The Twelfth International Conference on Learning Representations}}.
\newblock


\bibitem[Jain and Molino(2019)]%
        {jainenhancing}
\bibfield{author}{\bibinfo{person}{Ankit Jain} {and} \bibinfo{person}{Piero Molino}.} \bibinfo{year}{2019}\natexlab{}.
\newblock \showarticletitle{Enhancing Recommendations on Uber Eats with Graph Convolutional Networks}.
\newblock  (\bibinfo{year}{2019}).
\newblock


\bibitem[Joshi(2020)]%
        {joshi2020transformers}
\bibfield{author}{\bibinfo{person}{Chaitanya Joshi}.} \bibinfo{year}{2020}\natexlab{}.
\newblock \showarticletitle{Transformers are graph neural networks}.
\newblock \bibinfo{journal}{\emph{The Gradient}}  \bibinfo{volume}{12} (\bibinfo{year}{2020}), \bibinfo{pages}{17}.
\newblock


\bibitem[Kanatsoulis and Ribeiro(2024a)]%
        {kanatsoulis2024counting}
\bibfield{author}{\bibinfo{person}{Charilaos Kanatsoulis} {and} \bibinfo{person}{Alejandro Ribeiro}.} \bibinfo{year}{2024}\natexlab{a}.
\newblock \showarticletitle{Counting graph substructures with graph neural networks}. In \bibinfo{booktitle}{\emph{The twelfth international conference on learning representations}}.
\newblock


\bibitem[Kanatsoulis et~al\mbox{.}({[n.\,d.]})]%
        {kanatsoulis2025learning}
\bibfield{author}{\bibinfo{person}{Charilaos~I Kanatsoulis}, \bibinfo{person}{Evelyn Choi}, \bibinfo{person}{Stephanie Jegelka}, \bibinfo{person}{Jure Leskovec}, {and} \bibinfo{person}{Alejandro Ribeiro}.} \bibinfo{year}{[n.\,d.]}\natexlab{}.
\newblock \showarticletitle{Learning Efficient Positional Encodings with Graph Neural Networks}. In \bibinfo{booktitle}{\emph{The Thirteenth International Conference on Learning Representations}}.
\newblock


\bibitem[Kanatsoulis and Ribeiro(2024b)]%
        {kanatsoulis2024graph}
\bibfield{author}{\bibinfo{person}{Charilaos~I Kanatsoulis} {and} \bibinfo{person}{Alejandro Ribeiro}.} \bibinfo{year}{2024}\natexlab{b}.
\newblock \showarticletitle{Graph neural networks are more powerful than we think}. In \bibinfo{booktitle}{\emph{ICASSP 2024-2024 IEEE International Conference on Acoustics, Speech and Signal Processing (ICASSP)}}. IEEE, \bibinfo{pages}{7550--7554}.
\newblock


\bibitem[Kazemi et~al\mbox{.}(2019)]%
        {kazemi2019time2vec}
\bibfield{author}{\bibinfo{person}{Seyed~Mehran Kazemi}, \bibinfo{person}{Rishab Goel}, \bibinfo{person}{Sepehr Eghbali}, \bibinfo{person}{Janahan Ramanan}, \bibinfo{person}{Jaspreet Sahota}, \bibinfo{person}{Sanjay Thakur}, \bibinfo{person}{Stella Wu}, \bibinfo{person}{Cathal Smyth}, \bibinfo{person}{Pascal Poupart}, {and} \bibinfo{person}{Marcus Brubaker}.} \bibinfo{year}{2019}\natexlab{}.
\newblock \showarticletitle{Time2vec: Learning a vector representation of time}.
\newblock \bibinfo{journal}{\emph{arXiv preprint arXiv:1907.05321}} (\bibinfo{year}{2019}).
\newblock


\bibitem[Ke et~al\mbox{.}(2017)]%
        {ke2017lightgbm}
\bibfield{author}{\bibinfo{person}{Guolin Ke}, \bibinfo{person}{Qi Meng}, \bibinfo{person}{Thomas Finley}, \bibinfo{person}{Taifeng Wang}, \bibinfo{person}{Wei Chen}, \bibinfo{person}{Weidong Ma}, \bibinfo{person}{Qiwei Ye}, {and} \bibinfo{person}{Tie-Yan Liu}.} \bibinfo{year}{2017}\natexlab{}.
\newblock \showarticletitle{Lightgbm: A highly efficient gradient boosting decision tree}.
\newblock \bibinfo{journal}{\emph{Advances in neural information processing systems}}  \bibinfo{volume}{30} (\bibinfo{year}{2017}).
\newblock


\bibitem[Kipf and Welling(2016)]%
        {kipf2016semi}
\bibfield{author}{\bibinfo{person}{Thomas~N Kipf} {and} \bibinfo{person}{Max Welling}.} \bibinfo{year}{2016}\natexlab{}.
\newblock \showarticletitle{Semi-supervised classification with graph convolutional networks}.
\newblock \bibinfo{journal}{\emph{arXiv preprint arXiv:1609.02907}} (\bibinfo{year}{2016}).
\newblock


\bibitem[Kong et~al\mbox{.}(2023)]%
        {kong2023goat}
\bibfield{author}{\bibinfo{person}{Kezhi Kong}, \bibinfo{person}{Jiuhai Chen}, \bibinfo{person}{John Kirchenbauer}, \bibinfo{person}{Renkun Ni}, \bibinfo{person}{C~Bayan Bruss}, {and} \bibinfo{person}{Tom Goldstein}.} \bibinfo{year}{2023}\natexlab{}.
\newblock \showarticletitle{GOAT: A Global Transformer on Large-scale Graphs}. In \bibinfo{booktitle}{\emph{International Conference on Machine Learning}}.
\newblock


\bibitem[Kreuzer et~al\mbox{.}(2021)]%
        {kreuzer2021rethinking}
\bibfield{author}{\bibinfo{person}{Devin Kreuzer}, \bibinfo{person}{Dominique Beaini}, \bibinfo{person}{Will Hamilton}, \bibinfo{person}{Vincent L{\'e}tourneau}, {and} \bibinfo{person}{Prudencio Tossou}.} \bibinfo{year}{2021}\natexlab{}.
\newblock \showarticletitle{Rethinking graph transformers with spectral attention}.
\newblock \bibinfo{journal}{\emph{Advances in Neural Information Processing Systems}}  \bibinfo{volume}{34} (\bibinfo{year}{2021}), \bibinfo{pages}{21618--21629}.
\newblock


\bibitem[Lachi et~al\mbox{.}(2024)]%
        {lachiover}
\bibfield{author}{\bibinfo{person}{Veronica Lachi}, \bibinfo{person}{Antonio Longa}, \bibinfo{person}{Beatrice Bevilacqua}, \bibinfo{person}{Bruno Lepri}, \bibinfo{person}{Andrea Passerini}, {and} \bibinfo{person}{Bruno Ribeiro}.} \bibinfo{year}{2024}\natexlab{}.
\newblock \showarticletitle{Over 100x Speedup in Relational Deep Learning via Static GNNs and Tabular Distillation}.
\newblock  (\bibinfo{year}{2024}).
\newblock


\bibitem[LeCun et~al\mbox{.}(2015)]%
        {lecun2015deep}
\bibfield{author}{\bibinfo{person}{Yann LeCun}, \bibinfo{person}{Yoshua Bengio}, {and} \bibinfo{person}{Geoffrey Hinton}.} \bibinfo{year}{2015}\natexlab{}.
\newblock \showarticletitle{Deep learning}.
\newblock \bibinfo{journal}{\emph{nature}} \bibinfo{volume}{521}, \bibinfo{number}{7553} (\bibinfo{year}{2015}), \bibinfo{pages}{436--444}.
\newblock


\bibitem[Levie et~al\mbox{.}(2021)]%
        {levie2021transferability}
\bibfield{author}{\bibinfo{person}{Ron Levie}, \bibinfo{person}{Wei Huang}, \bibinfo{person}{Lorenzo Bucci}, \bibinfo{person}{Michael Bronstein}, {and} \bibinfo{person}{Gitta Kutyniok}.} \bibinfo{year}{2021}\natexlab{}.
\newblock \showarticletitle{Transferability of spectral graph convolutional neural networks}.
\newblock \bibinfo{journal}{\emph{Journal of Machine Learning Research}} \bibinfo{volume}{22}, \bibinfo{number}{272} (\bibinfo{year}{2021}), \bibinfo{pages}{1--59}.
\newblock


\bibitem[Li et~al\mbox{.}(2023)]%
        {li2023simplifying}
\bibfield{author}{\bibinfo{person}{Ce Li}, \bibinfo{person}{Rongpei Hong}, \bibinfo{person}{Xovee Xu}, \bibinfo{person}{Goce Trajcevski}, {and} \bibinfo{person}{Fan Zhou}.} \bibinfo{year}{2023}\natexlab{}.
\newblock \showarticletitle{Simplifying temporal heterogeneous network for continuous-time link prediction}. In \bibinfo{booktitle}{\emph{Proceedings of the 32nd ACM International Conference on Information and Knowledge Management}}. \bibinfo{pages}{1288--1297}.
\newblock


\bibitem[Lim et~al\mbox{.}(2022)]%
        {limsign}
\bibfield{author}{\bibinfo{person}{Derek Lim}, \bibinfo{person}{Joshua~David Robinson}, \bibinfo{person}{Lingxiao Zhao}, \bibinfo{person}{Tess Smidt}, \bibinfo{person}{Suvrit Sra}, \bibinfo{person}{Haggai Maron}, {and} \bibinfo{person}{Stefanie Jegelka}.} \bibinfo{year}{2022}\natexlab{}.
\newblock \showarticletitle{Sign and Basis Invariant Networks for Spectral Graph Representation Learning}. In \bibinfo{booktitle}{\emph{The Eleventh International Conference on Learning Representations}}.
\newblock


\bibitem[Loomis(2013)]%
        {loomis2013introduction}
\bibfield{author}{\bibinfo{person}{Lynn~H Loomis}.} \bibinfo{year}{2013}\natexlab{}.
\newblock \bibinfo{booktitle}{\emph{Introduction to abstract harmonic analysis}}.
\newblock \bibinfo{publisher}{Courier Corporation}.
\newblock


\bibitem[Luo and Li(2022)]%
        {luo2022neighborhood}
\bibfield{author}{\bibinfo{person}{Yuhong Luo} {and} \bibinfo{person}{Pan Li}.} \bibinfo{year}{2022}\natexlab{}.
\newblock \showarticletitle{Neighborhood-aware scalable temporal network representation learning}. In \bibinfo{booktitle}{\emph{Learning on Graphs Conference}}. PMLR, \bibinfo{pages}{1--1}.
\newblock


\bibitem[Mao et~al\mbox{.}(2024)]%
        {mao2024position}
\bibfield{author}{\bibinfo{person}{Haitao Mao}, \bibinfo{person}{Zhikai Chen}, \bibinfo{person}{Wenzhuo Tang}, \bibinfo{person}{Jianan Zhao}, \bibinfo{person}{Yao Ma}, \bibinfo{person}{Tong Zhao}, \bibinfo{person}{Neil Shah}, \bibinfo{person}{Mikhail Galkin}, {and} \bibinfo{person}{Jiliang Tang}.} \bibinfo{year}{2024}\natexlab{}.
\newblock \showarticletitle{Position: Graph foundation models are already here}. In \bibinfo{booktitle}{\emph{Forty-first International Conference on Machine Learning}}.
\newblock


\bibitem[Maron et~al\mbox{.}(2018)]%
        {maron2018invariant}
\bibfield{author}{\bibinfo{person}{Haggai Maron}, \bibinfo{person}{Heli Ben-Hamu}, \bibinfo{person}{Nadav Shamir}, {and} \bibinfo{person}{Yaron Lipman}.} \bibinfo{year}{2018}\natexlab{}.
\newblock \showarticletitle{Invariant and Equivariant Graph Networks}. In \bibinfo{booktitle}{\emph{International Conference on Learning Representations}}.
\newblock


\bibitem[Mialon et~al\mbox{.}(2021)]%
        {mialon2021graphit}
\bibfield{author}{\bibinfo{person}{Gr{\'e}goire Mialon}, \bibinfo{person}{Dexiong Chen}, \bibinfo{person}{Margot Selosse}, {and} \bibinfo{person}{Julien Mairal}.} \bibinfo{year}{2021}\natexlab{}.
\newblock \showarticletitle{Graphit: Encoding graph structure in transformers}.
\newblock \bibinfo{journal}{\emph{arXiv preprint arXiv:2106.05667}} (\bibinfo{year}{2021}).
\newblock


\bibitem[Morris et~al\mbox{.}(2019)]%
        {morris2019weisfeiler}
\bibfield{author}{\bibinfo{person}{Christopher Morris}, \bibinfo{person}{Martin Ritzert}, \bibinfo{person}{Matthias Fey}, \bibinfo{person}{William~L Hamilton}, \bibinfo{person}{Jan~Eric Lenssen}, \bibinfo{person}{Gaurav Rattan}, {and} \bibinfo{person}{Martin Grohe}.} \bibinfo{year}{2019}\natexlab{}.
\newblock \showarticletitle{Weisfeiler and leman go neural: higher-order graph neural networks}. In \bibinfo{booktitle}{\emph{Proceedings of the Thirty-Third AAAI Conference on Artificial Intelligence and Thirty-First Innovative Applications of Artificial Intelligence Conference and Ninth AAAI Symposium on Educational Advances in Artificial Intelligence}}. \bibinfo{pages}{4602--4609}.
\newblock


\bibitem[Motl and Schulte(2015)]%
        {motl2015ctu}
\bibfield{author}{\bibinfo{person}{Jan Motl} {and} \bibinfo{person}{Oliver Schulte}.} \bibinfo{year}{2015}\natexlab{}.
\newblock \showarticletitle{The CTU prague relational learning repository}.
\newblock \bibinfo{journal}{\emph{arXiv preprint arXiv:1511.03086}} (\bibinfo{year}{2015}).
\newblock


\bibitem[Pele{\v{s}}ka and {\v{S}}{\'\i}r(2024)]%
        {pelevska2024transformers}
\bibfield{author}{\bibinfo{person}{Jakub Pele{\v{s}}ka} {and} \bibinfo{person}{Gustav {\v{S}}{\'\i}r}.} \bibinfo{year}{2024}\natexlab{}.
\newblock \showarticletitle{Transformers Meet Relational Databases}.
\newblock \bibinfo{journal}{\emph{arXiv preprint arXiv:2412.05218}} (\bibinfo{year}{2024}).
\newblock


\bibitem[Radford et~al\mbox{.}(2018)]%
        {radford2018improving}
\bibfield{author}{\bibinfo{person}{Alec Radford}, \bibinfo{person}{Karthik Narasimhan}, \bibinfo{person}{Tim Salimans}, \bibinfo{person}{Ilya Sutskever}, {et~al\mbox{.}}} \bibinfo{year}{2018}\natexlab{}.
\newblock \bibinfo{title}{Improving language understanding by generative pre-training.(2018)}.
\newblock


\bibitem[Raffel et~al\mbox{.}(2020)]%
        {raffel2020exploring}
\bibfield{author}{\bibinfo{person}{Colin Raffel}, \bibinfo{person}{Noam Shazeer}, \bibinfo{person}{Adam Roberts}, \bibinfo{person}{Katherine Lee}, \bibinfo{person}{Sharan Narang}, \bibinfo{person}{Michael Matena}, \bibinfo{person}{Yanqi Zhou}, \bibinfo{person}{Wei Li}, {and} \bibinfo{person}{Peter~J Liu}.} \bibinfo{year}{2020}\natexlab{}.
\newblock \showarticletitle{Exploring the limits of transfer learning with a unified text-to-text transformer}.
\newblock \bibinfo{journal}{\emph{Journal of machine learning research}} \bibinfo{volume}{21}, \bibinfo{number}{140} (\bibinfo{year}{2020}), \bibinfo{pages}{1--67}.
\newblock


\bibitem[Ramp{\'a}{\v{s}}ek et~al\mbox{.}(2022)]%
        {rampavsek2022recipe}
\bibfield{author}{\bibinfo{person}{Ladislav Ramp{\'a}{\v{s}}ek}, \bibinfo{person}{Michael Galkin}, \bibinfo{person}{Vijay~Prakash Dwivedi}, \bibinfo{person}{Anh~Tuan Luu}, \bibinfo{person}{Guy Wolf}, {and} \bibinfo{person}{Dominique Beaini}.} \bibinfo{year}{2022}\natexlab{}.
\newblock \showarticletitle{Recipe for a general, powerful, scalable graph transformer}.
\newblock \bibinfo{journal}{\emph{Advances in Neural Information Processing Systems}}  \bibinfo{volume}{35} (\bibinfo{year}{2022}), \bibinfo{pages}{14501--14515}.
\newblock


\bibitem[Robinson et~al\mbox{.}(2024)]%
        {robinson2025relbench}
\bibfield{author}{\bibinfo{person}{Joshua Robinson}, \bibinfo{person}{Rishabh Ranjan}, \bibinfo{person}{Weihua Hu}, \bibinfo{person}{Kexin Huang}, \bibinfo{person}{Jiaqi Han}, \bibinfo{person}{Alejandro Dobles}, \bibinfo{person}{Matthias Fey}, \bibinfo{person}{Jan~Eric Lenssen}, \bibinfo{person}{Yiwen Yuan}, \bibinfo{person}{Zecheng Zhang}, {et~al\mbox{.}}} \bibinfo{year}{2024}\natexlab{}.
\newblock \showarticletitle{Relbench: A benchmark for deep learning on relational databases}.
\newblock \bibinfo{journal}{\emph{Advances in Neural Information Processing Systems}}  \bibinfo{volume}{37} (\bibinfo{year}{2024}), \bibinfo{pages}{21330--21341}.
\newblock


\bibitem[Rossi et~al\mbox{.}(2020)]%
        {rossi2020temporal}
\bibfield{author}{\bibinfo{person}{Emanuele Rossi}, \bibinfo{person}{Ben Chamberlain}, \bibinfo{person}{Fabrizio Frasca}, \bibinfo{person}{Davide Eynard}, \bibinfo{person}{Federico Monti}, {and} \bibinfo{person}{Michael Bronstein}.} \bibinfo{year}{2020}\natexlab{}.
\newblock \showarticletitle{Temporal graph networks for deep learning on dynamic graphs}.
\newblock \bibinfo{journal}{\emph{arXiv preprint arXiv:2006.10637}} (\bibinfo{year}{2020}).
\newblock


\bibitem[Ruiz et~al\mbox{.}(2020)]%
        {Ruiz2020}
\bibfield{author}{\bibinfo{person}{Luana Ruiz}, \bibinfo{person}{Luiz Chamon}, {and} \bibinfo{person}{Alejandro Ribeiro}.} \bibinfo{year}{2020}\natexlab{}.
\newblock \showarticletitle{Graphon Neural Networks and the Transferability of Graph Neural Networks}. In \bibinfo{booktitle}{\emph{Advances in Neural Information Processing Systems}}, Vol.~\bibinfo{volume}{33}. \bibinfo{pages}{1702--1712}.
\newblock


\bibitem[Rumelhart et~al\mbox{.}(1985)]%
        {rumelhart1985learning}
\bibfield{author}{\bibinfo{person}{David~E Rumelhart}, \bibinfo{person}{Geoffrey~E Hinton}, \bibinfo{person}{Ronald~J Williams}, {et~al\mbox{.}}} \bibinfo{year}{1985}\natexlab{}.
\newblock \bibinfo{title}{Learning internal representations by error propagation}.
\newblock


\bibitem[Schlichtkrull et~al\mbox{.}(2018)]%
        {schlichtkrull2018modeling}
\bibfield{author}{\bibinfo{person}{Michael Schlichtkrull}, \bibinfo{person}{Thomas~N Kipf}, \bibinfo{person}{Peter Bloem}, \bibinfo{person}{Rianne Van Den~Berg}, \bibinfo{person}{Ivan Titov}, {and} \bibinfo{person}{Max Welling}.} \bibinfo{year}{2018}\natexlab{}.
\newblock \showarticletitle{Modeling relational data with graph convolutional networks}. In \bibinfo{booktitle}{\emph{The semantic web: 15th international conference, ESWC 2018, Heraklion, Crete, Greece, June 3--7, 2018, proceedings 15}}. Springer, \bibinfo{pages}{593--607}.
\newblock


\bibitem[Shamsi et~al\mbox{.}(2024)]%
        {shamsi2024graphpulse}
\bibfield{author}{\bibinfo{person}{Kiarash Shamsi}, \bibinfo{person}{Farimah Poursafaei}, \bibinfo{person}{Shenyang Huang}, \bibinfo{person}{Bao Tran~Gia Ngo}, \bibinfo{person}{Baris Coskunuzer}, {and} \bibinfo{person}{Cuneyt~Gurcan Akcora}.} \bibinfo{year}{2024}\natexlab{}.
\newblock \showarticletitle{GraphPulse: Topological representations for temporal graph property prediction}. In \bibinfo{booktitle}{\emph{The Twelfth International Conference on Learning Representations}}.
\newblock


\bibitem[Shirzad et~al\mbox{.}(2023)]%
        {shirzad2023exphormer}
\bibfield{author}{\bibinfo{person}{Hamed Shirzad}, \bibinfo{person}{Ameya Velingker}, \bibinfo{person}{Balaji Venkatachalam}, \bibinfo{person}{Danica~J Sutherland}, {and} \bibinfo{person}{Ali~Kemal Sinop}.} \bibinfo{year}{2023}\natexlab{}.
\newblock \showarticletitle{Exphormer: Sparse transformers for graphs}.
\newblock \bibinfo{journal}{\emph{arXiv preprint arXiv:2303.06147}} (\bibinfo{year}{2023}).
\newblock


\bibitem[Sypetkowski et~al\mbox{.}(2025)]%
        {sypetkowski2025scalability}
\bibfield{author}{\bibinfo{person}{Maciej Sypetkowski}, \bibinfo{person}{Frederik Wenkel}, \bibinfo{person}{Farimah Poursafaei}, \bibinfo{person}{Nia Dickson}, \bibinfo{person}{Karush Suri}, \bibinfo{person}{Philip Fradkin}, {and} \bibinfo{person}{Dominique Beaini}.} \bibinfo{year}{2025}\natexlab{}.
\newblock \showarticletitle{On the scalability of gnns for molecular graphs}.
\newblock \bibinfo{journal}{\emph{Advances in Neural Information Processing Systems}}  \bibinfo{volume}{37} (\bibinfo{year}{2025}), \bibinfo{pages}{19870--19906}.
\newblock


\bibitem[Tay et~al\mbox{.}(2022)]%
        {tay2022ul2}
\bibfield{author}{\bibinfo{person}{Yi Tay}, \bibinfo{person}{Mostafa Dehghani}, \bibinfo{person}{Vinh~Q Tran}, \bibinfo{person}{Xavier Garcia}, \bibinfo{person}{Jason Wei}, \bibinfo{person}{Xuezhi Wang}, \bibinfo{person}{Hyung~Won Chung}, \bibinfo{person}{Siamak Shakeri}, \bibinfo{person}{Dara Bahri}, \bibinfo{person}{Tal Schuster}, {et~al\mbox{.}}} \bibinfo{year}{2022}\natexlab{}.
\newblock \showarticletitle{Ul2: Unifying language learning paradigms}.
\newblock \bibinfo{journal}{\emph{arXiv preprint arXiv:2205.05131}} (\bibinfo{year}{2022}).
\newblock


\bibitem[Team et~al\mbox{.}(2023)]%
        {team2023gemini}
\bibfield{author}{\bibinfo{person}{Gemini Team}, \bibinfo{person}{Rohan Anil}, \bibinfo{person}{Sebastian Borgeaud}, \bibinfo{person}{Jean-Baptiste Alayrac}, \bibinfo{person}{Jiahui Yu}, \bibinfo{person}{Radu Soricut}, \bibinfo{person}{Johan Schalkwyk}, \bibinfo{person}{Andrew~M Dai}, \bibinfo{person}{Anja Hauth}, \bibinfo{person}{Katie Millican}, {et~al\mbox{.}}} \bibinfo{year}{2023}\natexlab{}.
\newblock \showarticletitle{Gemini: a family of highly capable multimodal models}.
\newblock \bibinfo{journal}{\emph{arXiv preprint arXiv:2312.11805}} (\bibinfo{year}{2023}).
\newblock


\bibitem[Team et~al\mbox{.}(2024a)]%
        {team2024gemma}
\bibfield{author}{\bibinfo{person}{Gemma Team}, \bibinfo{person}{Thomas Mesnard}, \bibinfo{person}{Cassidy Hardin}, \bibinfo{person}{Robert Dadashi}, \bibinfo{person}{Surya Bhupatiraju}, \bibinfo{person}{Shreya Pathak}, \bibinfo{person}{Laurent Sifre}, \bibinfo{person}{Morgane Rivi{\`e}re}, \bibinfo{person}{Mihir~Sanjay Kale}, \bibinfo{person}{Juliette Love}, {et~al\mbox{.}}} \bibinfo{year}{2024}\natexlab{a}.
\newblock \showarticletitle{Gemma: Open models based on gemini research and technology}.
\newblock \bibinfo{journal}{\emph{arXiv preprint arXiv:2403.08295}} (\bibinfo{year}{2024}).
\newblock


\bibitem[Team et~al\mbox{.}(2024b)]%
        {team2024reka}
\bibfield{author}{\bibinfo{person}{Reka Team}, \bibinfo{person}{Aitor Ormazabal}, \bibinfo{person}{Che Zheng}, \bibinfo{person}{Cyprien de~Masson d'Autume}, \bibinfo{person}{Dani Yogatama}, \bibinfo{person}{Deyu Fu}, \bibinfo{person}{Donovan Ong}, \bibinfo{person}{Eric Chen}, \bibinfo{person}{Eugenie Lamprecht}, \bibinfo{person}{Hai Pham}, {et~al\mbox{.}}} \bibinfo{year}{2024}\natexlab{b}.
\newblock \showarticletitle{Reka core, flash, and edge: A series of powerful multimodal language models}.
\newblock \bibinfo{journal}{\emph{arXiv preprint arXiv:2404.12387}} (\bibinfo{year}{2024}).
\newblock


\bibitem[Vaswani et~al\mbox{.}(2017)]%
        {vaswani2017attention}
\bibfield{author}{\bibinfo{person}{Ashish Vaswani}, \bibinfo{person}{Noam Shazeer}, \bibinfo{person}{Niki Parmar}, \bibinfo{person}{Jakob Uszkoreit}, \bibinfo{person}{Llion Jones}, \bibinfo{person}{Aidan~N Gomez}, \bibinfo{person}{{\L}ukasz Kaiser}, {and} \bibinfo{person}{Illia Polosukhin}.} \bibinfo{year}{2017}\natexlab{}.
\newblock \showarticletitle{Attention is all you need}.
\newblock \bibinfo{journal}{\emph{Advances in Neural Information Processing Systems}}  \bibinfo{volume}{30} (\bibinfo{year}{2017}).
\newblock


\bibitem[Velickovic et~al\mbox{.}(2017)]%
        {velickovic2017graph}
\bibfield{author}{\bibinfo{person}{Petar Velickovic}, \bibinfo{person}{Guillem Cucurull}, \bibinfo{person}{Arantxa Casanova}, \bibinfo{person}{Adriana Romero}, \bibinfo{person}{Pietro Lio}, \bibinfo{person}{Yoshua Bengio}, {et~al\mbox{.}}} \bibinfo{year}{2017}\natexlab{}.
\newblock \showarticletitle{Graph attention networks}.
\newblock \bibinfo{journal}{\emph{stat}} \bibinfo{volume}{1050}, \bibinfo{number}{20} (\bibinfo{year}{2017}), \bibinfo{pages}{10--48550}.
\newblock


\bibitem[Wang et~al\mbox{.}(2021)]%
        {wang2021tcl}
\bibfield{author}{\bibinfo{person}{Lu Wang}, \bibinfo{person}{Xiaofu Chang}, \bibinfo{person}{Shuang Li}, \bibinfo{person}{Yunfei Chu}, \bibinfo{person}{Hui Li}, \bibinfo{person}{Wei Zhang}, \bibinfo{person}{Xiaofeng He}, \bibinfo{person}{Le Song}, \bibinfo{person}{Jingren Zhou}, {and} \bibinfo{person}{Hongxia Yang}.} \bibinfo{year}{2021}\natexlab{}.
\newblock \showarticletitle{Tcl: Transformer-based dynamic graph modelling via contrastive learning}.
\newblock \bibinfo{journal}{\emph{arXiv preprint arXiv:2105.07944}} (\bibinfo{year}{2021}).
\newblock


\bibitem[Wang et~al\mbox{.}(2019)]%
        {wang2019heterogeneous}
\bibfield{author}{\bibinfo{person}{Xiao Wang}, \bibinfo{person}{Houye Ji}, \bibinfo{person}{Chuan Shi}, \bibinfo{person}{Bai Wang}, \bibinfo{person}{Yanfang Ye}, \bibinfo{person}{Peng Cui}, {and} \bibinfo{person}{Philip~S Yu}.} \bibinfo{year}{2019}\natexlab{}.
\newblock \showarticletitle{Heterogeneous graph attention network}. In \bibinfo{booktitle}{\emph{The world wide web conference}}. \bibinfo{pages}{2022--2032}.
\newblock


\bibitem[Wang et~al\mbox{.}(2020)]%
        {wang2020inductive}
\bibfield{author}{\bibinfo{person}{Yanbang Wang}, \bibinfo{person}{Yen-Yu Chang}, \bibinfo{person}{Yunyu Liu}, \bibinfo{person}{Jure Leskovec}, {and} \bibinfo{person}{Pan Li}.} \bibinfo{year}{2020}\natexlab{}.
\newblock \showarticletitle{Inductive Representation Learning in Temporal Networks via Causal Anonymous Walks}. In \bibinfo{booktitle}{\emph{International Conference on Learning Representations}}.
\newblock


\bibitem[Wydmuch et~al\mbox{.}(2024)]%
        {wydmuch2024tackling}
\bibfield{author}{\bibinfo{person}{Marek Wydmuch}, \bibinfo{person}{{\L}ukasz Borchmann}, {and} \bibinfo{person}{Filip Grali{\'n}ski}.} \bibinfo{year}{2024}\natexlab{}.
\newblock \showarticletitle{Tackling prediction tasks in relational databases with LLMs}.
\newblock \bibinfo{journal}{\emph{arXiv preprint arXiv:2411.11829}} (\bibinfo{year}{2024}).
\newblock


\bibitem[Xu et~al\mbox{.}(2020)]%
        {xu2020inductive}
\bibfield{author}{\bibinfo{person}{Da Xu}, \bibinfo{person}{Chuanwei Ruan}, \bibinfo{person}{Evren Korpeoglu}, \bibinfo{person}{Sushant Kumar}, {and} \bibinfo{person}{Kannan Achan}.} \bibinfo{year}{2020}\natexlab{}.
\newblock \showarticletitle{Inductive representation learning on temporal graphs}.
\newblock \bibinfo{journal}{\emph{arXiv preprint arXiv:2002.07962}} (\bibinfo{year}{2020}).
\newblock


\bibitem[Xu et~al\mbox{.}(2019)]%
        {xu2018powerful}
\bibfield{author}{\bibinfo{person}{Keyulu Xu}, \bibinfo{person}{Weihua Hu}, \bibinfo{person}{Jure Leskovec}, {and} \bibinfo{person}{Stefanie Jegelka}.} \bibinfo{year}{2019}\natexlab{}.
\newblock \showarticletitle{How Powerful are Graph Neural Networks?}. In \bibinfo{booktitle}{\emph{International Conference on Learning Representations}}.
\newblock
\urldef\tempurl%
\url{https://openreview.net/forum?id=ryGs6iA5Km}
\showURL{%
\tempurl}


\bibitem[Yang(2019)]%
        {yang2019aligraph}
\bibfield{author}{\bibinfo{person}{Hongxia Yang}.} \bibinfo{year}{2019}\natexlab{}.
\newblock \showarticletitle{Aligraph: A comprehensive graph neural network platform}. In \bibinfo{booktitle}{\emph{Proceedings of the 25th ACM SIGKDD international conference on knowledge discovery \& data mining}}. \bibinfo{pages}{3165--3166}.
\newblock


\bibitem[Ying et~al\mbox{.}(2021)]%
        {ying2021transformers}
\bibfield{author}{\bibinfo{person}{Chengxuan Ying}, \bibinfo{person}{Tianle Cai}, \bibinfo{person}{Shengjie Luo}, \bibinfo{person}{Shuxin Zheng}, \bibinfo{person}{Guolin Ke}, \bibinfo{person}{Di He}, \bibinfo{person}{Yanming Shen}, {and} \bibinfo{person}{Tie-Yan Liu}.} \bibinfo{year}{2021}\natexlab{}.
\newblock \showarticletitle{Do transformers really perform badly for graph representation?}
\newblock \bibinfo{journal}{\emph{Advances in Neural Information Processing Systems}}  \bibinfo{volume}{34} (\bibinfo{year}{2021}), \bibinfo{pages}{28877--28888}.
\newblock


\bibitem[Ying et~al\mbox{.}(2018)]%
        {ying2018graph}
\bibfield{author}{\bibinfo{person}{Rex Ying}, \bibinfo{person}{Ruining He}, \bibinfo{person}{Kaifeng Chen}, \bibinfo{person}{Pong Eksombatchai}, \bibinfo{person}{William~L Hamilton}, {and} \bibinfo{person}{Jure Leskovec}.} \bibinfo{year}{2018}\natexlab{}.
\newblock \showarticletitle{Graph convolutional neural networks for web-scale recommender systems}. In \bibinfo{booktitle}{\emph{Proceedings of the 24th ACM SIGKDD international conference on knowledge discovery \& data mining}}. \bibinfo{pages}{974--983}.
\newblock


\bibitem[You et~al\mbox{.}(2022)]%
        {you2022roland}
\bibfield{author}{\bibinfo{person}{Jiaxuan You}, \bibinfo{person}{Tianyu Du}, {and} \bibinfo{person}{Jure Leskovec}.} \bibinfo{year}{2022}\natexlab{}.
\newblock \showarticletitle{ROLAND: graph learning framework for dynamic graphs}. In \bibinfo{booktitle}{\emph{Proceedings of the 28th ACM SIGKDD conference on knowledge discovery and data mining}}. \bibinfo{pages}{2358--2366}.
\newblock


\bibitem[Yu et~al\mbox{.}(2023)]%
        {yu2023towards}
\bibfield{author}{\bibinfo{person}{Le Yu}, \bibinfo{person}{Leilei Sun}, \bibinfo{person}{Bowen Du}, {and} \bibinfo{person}{Weifeng Lv}.} \bibinfo{year}{2023}\natexlab{}.
\newblock \showarticletitle{Towards better dynamic graph learning: New architecture and unified library}.
\newblock \bibinfo{journal}{\emph{Advances in Neural Information Processing Systems}}  \bibinfo{volume}{36} (\bibinfo{year}{2023}), \bibinfo{pages}{67686--67700}.
\newblock


\bibitem[Yuan et~al\mbox{.}(2024)]%
        {yuan2024contextgnn}
\bibfield{author}{\bibinfo{person}{Yiwen Yuan}, \bibinfo{person}{Zecheng Zhang}, \bibinfo{person}{Xinwei He}, \bibinfo{person}{Akihiro Nitta}, \bibinfo{person}{Weihua Hu}, \bibinfo{person}{Dong Wang}, \bibinfo{person}{Manan Shah}, \bibinfo{person}{Shenyang Huang}, \bibinfo{person}{Bla{\v{z}} Stojanovi{\v{c}}}, \bibinfo{person}{Alan Krumholz}, {et~al\mbox{.}}} \bibinfo{year}{2024}\natexlab{}.
\newblock \showarticletitle{ContextGNN: Beyond Two-Tower Recommendation Systems}.
\newblock \bibinfo{journal}{\emph{arXiv preprint arXiv:2411.19513}} (\bibinfo{year}{2024}).
\newblock


\bibitem[Zahradn{\'\i}k et~al\mbox{.}(2023)]%
        {zahradnik2023deep}
\bibfield{author}{\bibinfo{person}{Luk{\'a}{\v{s}} Zahradn{\'\i}k}, \bibinfo{person}{Jan Neumann}, {and} \bibinfo{person}{Gustav {\v{S}}{\'\i}r}.} \bibinfo{year}{2023}\natexlab{}.
\newblock \showarticletitle{A deep learning blueprint for relational databases}. In \bibinfo{booktitle}{\emph{NeurIPS 2023 Second Table Representation Learning Workshop}}.
\newblock


\bibitem[Zhang et~al\mbox{.}(2023)]%
        {zhang2023rethinking}
\bibfield{author}{\bibinfo{person}{Bohang Zhang}, \bibinfo{person}{Shengjie Luo}, \bibinfo{person}{Liwei Wang}, {and} \bibinfo{person}{Di He}.} \bibinfo{year}{2023}\natexlab{}.
\newblock \showarticletitle{Rethinking the expressive power of gnns via graph biconnectivity}.
\newblock \bibinfo{journal}{\emph{arXiv preprint arXiv:2301.09505}} (\bibinfo{year}{2023}).
\newblock


\bibitem[Zhang et~al\mbox{.}(2022)]%
        {zhang2022hierarchical}
\bibfield{author}{\bibinfo{person}{Zaixi Zhang}, \bibinfo{person}{Qi Liu}, \bibinfo{person}{Qingyong Hu}, {and} \bibinfo{person}{Chee-Kong Lee}.} \bibinfo{year}{2022}\natexlab{}.
\newblock \showarticletitle{Hierarchical graph transformer with adaptive node sampling}.
\newblock \bibinfo{journal}{\emph{Advances in Neural Information Processing Systems}}  \bibinfo{volume}{35} (\bibinfo{year}{2022}), \bibinfo{pages}{21171--21183}.
\newblock


\bibitem[Zhao et~al\mbox{.}(2021)]%
        {zhao2021gophormer}
\bibfield{author}{\bibinfo{person}{Jianan Zhao}, \bibinfo{person}{Chaozhuo Li}, \bibinfo{person}{Qianlong Wen}, \bibinfo{person}{Yiqi Wang}, \bibinfo{person}{Yuming Liu}, \bibinfo{person}{Hao Sun}, \bibinfo{person}{Xing Xie}, {and} \bibinfo{person}{Yanfang Ye}.} \bibinfo{year}{2021}\natexlab{}.
\newblock \showarticletitle{Gophormer: Ego-graph transformer for node classification}.
\newblock \bibinfo{journal}{\emph{arXiv preprint arXiv:2110.13094}} (\bibinfo{year}{2021}).
\newblock


\bibitem[Zhao et~al\mbox{.}(2025)]%
        {zhao2025gigl}
\bibfield{author}{\bibinfo{person}{Tong Zhao}, \bibinfo{person}{Yozen Liu}, \bibinfo{person}{Matthew Kolodner}, \bibinfo{person}{Kyle Montemayor}, \bibinfo{person}{Elham Ghazizadeh}, \bibinfo{person}{Ankit Batra}, \bibinfo{person}{Zihao Fan}, \bibinfo{person}{Xiaobin Gao}, \bibinfo{person}{Xuan Guo}, \bibinfo{person}{Jiwen Ren}, {et~al\mbox{.}}} \bibinfo{year}{2025}\natexlab{}.
\newblock \showarticletitle{GiGL: Large-Scale Graph Neural Networks at Snapchat}.
\newblock \bibinfo{journal}{\emph{arXiv preprint arXiv:2502.15054}} (\bibinfo{year}{2025}).
\newblock


\bibitem[Zhu et~al\mbox{.}(2023a)]%
        {zhu2023xtab}
\bibfield{author}{\bibinfo{person}{Bingzhao Zhu}, \bibinfo{person}{Xingjian Shi}, \bibinfo{person}{Nick Erickson}, \bibinfo{person}{Mu Li}, \bibinfo{person}{George Karypis}, {and} \bibinfo{person}{Mahsa Shoaran}.} \bibinfo{year}{2023}\natexlab{a}.
\newblock \showarticletitle{XTab: Cross-table Pretraining for Tabular Transformers}. In \bibinfo{booktitle}{\emph{ICML'23: Proceedings of the 40th International Conference on Machine Learning}}. JMLR. org.
\newblock


\bibitem[Zhu et~al\mbox{.}(2023b)]%
        {zhu2023hierarchical}
\bibfield{author}{\bibinfo{person}{Wenhao Zhu}, \bibinfo{person}{Tianyu Wen}, \bibinfo{person}{Guojie Song}, \bibinfo{person}{Xiaojun Ma}, {and} \bibinfo{person}{Liang Wang}.} \bibinfo{year}{2023}\natexlab{b}.
\newblock \showarticletitle{Hierarchical Transformer for Scalable Graph Learning}.
\newblock \bibinfo{journal}{\emph{arXiv preprint arXiv:2305.02866}} (\bibinfo{year}{2023}).
\newblock


\end{thebibliography}

\appendix

\end{document}